\documentclass[letterpaper]{article} 
\usepackage{aaai2026}  
\usepackage{times}  
\usepackage{helvet}  
\usepackage{courier}  
\usepackage[hyphens]{url}  
\usepackage{graphicx} 
\usepackage{natbib}  
\usepackage{amsmath}
\usepackage{amssymb}
\usepackage{amsthm}
\usepackage{caption} 
\usepackage{algorithm}
\usepackage{xcolor}
\usepackage{caption}  
\usepackage{booktabs}
\usepackage{algpseudocode}
\usepackage{multirow}
\usepackage{subcaption}
\newtheorem{theorem}{Theorem}
\usepackage{newfloat}
\usepackage{listings}
\DeclareCaptionStyle{ruled}{labelfont=normalfont,labelsep=colon,strut=off} 
\floatstyle{ruled}
\newfloat{listing}{tb}{lst}{}
\floatname{listing}{Listing}
\title{Decoupling Understanding from Reasoning via Problem Space Mapping for Small-Scale Model Reasoning}
\author {
    Li Wang\textsuperscript{\rm 1},
    Changhao Zhang\textsuperscript{\rm 2},
    Zengqi Xiu\textsuperscript{\rm 1},
    Kai Lu\textsuperscript{\rm 3,\rm 4},
    Xin Yu\textsuperscript{\rm 1},
    Kui Zhang\textsuperscript{\rm 1},
    Wenjun Wu\textsuperscript{\rm 1,\rm 5,\rm 6}\thanks{Corresponding author.}
}
\affiliations {
    \textsuperscript{\rm 1}Beihang University, Beijing, China\\
    \textsuperscript{\rm 2}UCL Hawkes Institute and Department of Medical Physics and Biomedical
    Engineering, University College London, UK\\
    \textsuperscript{\rm 3}State Key Laboratory of Multimodal Artificial Intelligence Systems, Institute of Automation, Chinese Academy of Sciences, Beijing, China\\
    \textsuperscript{\rm 4}School of Artificial Intelligence, University of Chinese Academy of Sciences, Beijing, China\\
    \textsuperscript{\rm 5}Hangzhou International Innovation Institute, Beihang University, Hangzhou, China\\
    \textsuperscript{\rm 6}Beijing Advanced Innovation Center for Future Blockchain and Privacy Computing, Beihang University, Beijing, China\\
    {\{wangli\_42, xiuzengqi, nlsdeyuxin, zhangkui, wwj09315\}}@buaa.edu.cn, changhao.zhang.24@ucl.ac.uk, lukai24@mails.ucas.ac.cn 
    
}

\usepackage{bibentry}

\begin{document}

\maketitle

\begin{abstract}
Despite recent advances in the reasoning capabilities of Large Language Models (LLMs), improving the reasoning ability of Small Language Models (SLMs, e.g., up to 1.5B parameters) remains challenging.  A key obstacle lies in the complexity and variability of natural language: essentially equivalent problems often appear in diverse surface forms, often obscured by redundant or distracting details. This imposes a dual burden on SLMs: they must first extract the core problem from complex linguistic input, and then perform reasoning based on that understanding. The resulting vast and noisy problem space hinders optimization, particularly for models with limited capacity. To address this, we propose a new framework that decouples understanding from reasoning by mapping natural language problems into a canonical problem space-a semantically simplified yet expressive domain. This enables SLMs to focus on reasoning over standardized inputs, free from linguistic variability. Within this framework, we introduce DURIT (Decoupled Understanding from Reasoning via Iterative Training), a three-step algorithm that iteratively: (1) mapping natural language problems via reinforcement learning, (2) aligns reasoning trajectories through self-distillation, and (3) trains reasoning policies in the problem space. The mapper and reasoner are co-trained in an alternating loop throughout this process. Experiments show that DURIT substantially improves SLMs' performance on both in-domain and out-of-domain mathematical and logical reasoning tasks. Beyond improving reasoning capabilities, DURIT also improves the robustness of reasoning, validating decoupling understanding from reasoning as an effective strategy for strengthening SLMs.
\end{abstract}

\begin{links}
    \link{Code}{https://github.com/monster476/DURIT}
\end{links}

\section{Introduction}

Large Language Models (LLMs) \cite{yang2025qwen3} have demonstrated remarkable advances in reasoning capabilities \cite{bi2025forestofthoughtscalingtesttimecompute, luo2025wizardmathempoweringmathematicalreasoning, wen2024codeplan}. However, most existing research has primarily focused on relatively large models \cite{guan2025rstarmathsmallllmsmaster, li2025policyguidedtreesearch, shen2025satorireinforcementlearningchainofactionthought}, while the reasoning abilities of Small Language Models (SLMs, e.g., $\leq$ 1.5B) remain not fully explored. Despite their limited capacity, SLMs hold significant promise in edge-deployed scenarios and latency-sensitive applications due to their compact size and fast inference \cite{sun2020mobilebertcompacttaskagnosticbert, xu2024ondevicelanguagemodelscomprehensive}. Nevertheless, enhancing their reasoning capabilities remains a significant challenge due to their limited parameter capacity.

Recent efforts to improve LLM reasoning have focused on enhancing Chains of Thought (CoT)~\cite{wei2022chain}, using techniques like search-based reasoning~\cite{li2025policyguidedtreesearch, guan2025rstarmathsmallllmsmaster} and error correction~\cite{ma2025s2rteachingllmsselfverify, yang2025supercorrectadvancingsmallllm}. However, due to limited capacity, SLMs struggle to generate complex reasoning traces, making such approaches less effective. Knowledge Distillation (KD) is a common strategy for improving SLMs by transferring reasoning abilities from larger teacher LLMs via teacher-generated traces (e.g., CoT) or token-level supervision. However, distribution and capacity mismatches between teacher and student models pose challenges for both data and teacher selection. KD heavily depends on diverse, high-quality data~\cite{gu2025miniplmknowledgedistillationpretraining}: overly simple examples may cause overfitting to shallow patterns~\cite{shumailov2024ai}, while complex CoT traces may exceed the capacity of SLMs and hinder learning~\cite{li2025smallmodelsstrugglelearn}. Teacher-student mismatches can further degrade performance~\cite{Cho_2019_ICCV, chen2025unveilingkeyfactorsdistilling}, underscoring the challenge of distilling high-quality reasoning into SLMs.

Unlike KD, reinforcement learning (RL) enables models to autonomously explore solutions, often yielding stronger generalization~\cite{chu2025sftmemorizesrlgeneralizes, huan2025doesmathreasoningimprove}. The strong performance of DeepSeek-R1~\cite{shao2024deepseekmathpushinglimitsmathematical} further underscores RL’s potential in enhancing the reasoning capabilities of LLMs. However, SLMs face unique challenges: they must comprehend the semantic complexity of natural language problems and perform multi-step reasoning despite limited capacity. The vast state space induced by natural language severely limits the efficiency of RL. Even superficial variations in problem phrasing can mislead models~\cite{mirzadeh2024gsmsymbolicunderstandinglimitationsmathematical, liu2025cogmathassessingllmsauthentic}, which often rely on shallow heuristics rather than genuine understanding. This suggests that models may fail to grasp the essence of the problems and are easily distracted by surface-level linguistic variations. In contrast, humans readily generalize across diverse surface forms once they grasp the essence of the problems. This contrast raises a key question: how can models acquire such essential understanding and generalize in a human-like manner? We address this by proposing a new perspective—rather than reasoning directly over the high-dimensional, noisy space of natural language, we first map problems into a lower-dimensional, standardized problem space. This transformation reduces spurious variability and constrains the search space by clustering essentially similar problems into more representative and canonical forms. As a result, it compresses the state space, highlights the essence of the problem, and alleviates the burden of superficial language understanding, thereby improving exploration efficiency. Crucially, our approach is orthogonal to existing CoT-based methods: problem space transformation acts as a front-end normalization layer, enabling more effective and robust downstream reasoning. 

In this paper, we propose a general framework that maps natural language problems into a more abstract, low-dimensional, and semantically canonical space, effectively reducing the complexity of the original problem space. Within this space, models can learn and reason more efficiently. We instantiate this framework with a novel three-step alternating training algorithm: (1) a problem-space mapper is trained using RL and implicit templates to map natural language problems into standardized, low-dimensional forms; (2) self-distillation transfers this mapping capability into a SLM; and (3) a reasoning model is trained via RL to operate within the problem space. The mapper and the SLM are optimized in an alternating fashion, enabling iterative improvement in reasoning ability. To validate the effectiveness of DURIT, we conduct comprehensive empirical studies using models from the LLaMA~\cite{grattafiori2024llama} and Qwen~\cite{yang2025qwen3} families, with parameter sizes ranging from 0.5B to 1.5B. Even when trained solely on the GSM8K~\cite{cobbe2021trainingverifierssolvemath} dataset, DURIT achieves significant gains across a range of in-domain and out-of-domain datasets, including those focused on mathematical and logical reasoning, and demonstrates strong generalization capabilities. Unlike traditional CoT-based approaches, DURIT introduces a paradigm that enhances reasoning by compressing the problem space. Our main contributions are summarized as follows:
\begin{itemize}
    \item We propose a general framework that maps natural language problems into a standardized, low-dimensional space, reducing the effective state space and improving exploration and sample efficiency.
    \item We introduce DURIT, a three-step alternating training algorithm that decouples understanding from reasoning and progressively enhances the reasoning ability and robustness of SLMs through iterative co-training of a problem-space mapper and a reasoning model.
    \item Experiments show that DURIT yields substantial performance gains on both mathematical and logical reasoning tasks, across in-domain and out-of-domain settings, even with limited training data. In addition to improved accuracy, DURIT enhances reasoning robustness, indicating a deeper grasp of the problem’s underlying essence and improved generalization across varied formulations.
\end{itemize}

\section{Related Work}
\subsection{Prompt Optimization}
Prompt optimization improves test-time performance by refining LLM inputs. Some approaches use paraphrasing~\cite{NEURIPS2021_e4d2b6e6, deng2024rephraserespondletlarge}, while others apply RL to explore prompt formats more effectively~\cite{deng2022rlpromptoptimizingdiscretetext, zhang2022temperatesttimepromptingreinforcement}. PRewrite~\cite{kong-etal-2024-prewrite} trains an LLM via PPO~\cite{schulman2017proximalpolicyoptimizationalgorithms} using response accuracy as reward, but incurs high inference cost due to LLM-based prompt generation. AbstRaL~\cite{gao2025abstralaugmentingllmsreasoning} improves reasoning robustness by abstracting problems into symbolic forms and delegating reasoning to external toolchains. Unlike prior work, our goal is to eliminate reliance on external tools and enable reasoning entirely within the natural language space. To achieve this, we train a problem space mapper via RL and distill its transformation behavior into the SLM, leading to improved reasoning performance and robustness.

\subsection{Knowledge Distillation}
Knowledge distillation transfers knowledge from a large teacher to a smaller student and can be divided into offline and online paradigms. Offline KD uses teacher-generated data. Std-CoT~\cite{magister-etal-2023-teaching} fine-tunes students on CoT demonstrations, while NesyCD~\cite{Liao_He_Xu_Zhang_Liu_Zhao_2025} distills general capabilities and incorporates external knowledge. Online KD requires the teacher to provide token-level supervision during inference. Vanilla-KD~\cite{NEURIPS2024_48229913} distills hidden states and output probabilities, BOND~\cite{sessa2024bond} employs self-distillation based on the model’s best responses, and STaR~\cite{zelikman2024star} fine-tunes on self-generated CoT traces with correct final answers to improve performance. In contrast to prior work, our method focuses on self-distillation to transfer knowledge internally, enabling the model to generalize its learned capabilities to unfamiliar tasks.

\subsection{Reinforcement Learning for LLM Reasoning}
Reinforcement Learning has proven effective in enhancing the capabilities of large language models. Reinforcement learning from human feedback (RLHF) \cite{bai2022training, NEURIPS2022_b1efde53} is now a standard approach for aligning model outputs with human preferences. Recent work such as DeepSeek-R1 \cite{shao2024deepseekmathpushinglimitsmathematical} and Kimi K1.5 \cite{team2025kimi} shows that techniques like GRPO can significantly boost reasoning ability, highlighting the promise of RL with verifiable rewards (RLVR). Building on this, many studies have proposed further refinements \cite{yu2025dapo, team2025kimi, liu2025understandingr1zeroliketrainingcritical}. However, the vast and complex state space of natural language poses a major challenge to efficient exploration. To address this, we propose a problem space mapping that projects the original space into a lower-dimensional, more organized representation, thereby improving RL efficiency.

\section{Decoupling Language Understanding From Reasoning}
The inherent complexity of natural language presents a dual challenge for SLMs: interpreting subtle semantic nuances and performing reasoning, both constrained by limited model capacity. To address this, we propose a general framework that decouples understanding from reasoning. At its core is the notion of a problem space—a standardized, low-dimensional representation that abstracts surface variability while preserving essential semantics. By mapping fundamentally similar questions to nearby representations, the problem space reduces input complexity and offers a more interpretable and learning-efficient interface for downstream reasoning. As shown in the Appendix F\ref{appendix:case_study}, standardizing complex questions mitigates misinterpretation and improves reasoning accuracy. Formally, let $\mathcal{Q}$ denote the space of natural language questions, and let $\mathcal{P} \subset \mathcal{L}$ be a finite set of canonical forms drawn from the natural language space $\mathcal{L}$. We define a mapping $f: \mathcal{Q} \rightarrow \mathcal{P}$ that assigns each question $q \in \mathcal{Q}$ to a canonical representation $p = f(q) \in \mathcal{P}$. The construction of $\mathcal{P}$ and $f$ is guided by the objective:
\begin{equation}
\begin{aligned}
& \max_{f} \quad \mathbb{E}_{q \sim \mathcal{Q}} \left[ \operatorname{Acc}(f(q); \theta) \right] \\
& \text{s.t.} \quad 
\begin{cases}
\dim(\mathcal{P}) < \dim(\mathcal{Q}), \\
\operatorname{Dist}(f(q_1), f(q_2)) \leq \epsilon, \quad \forall (q_1, q_2) \in \mathcal{S},
\end{cases}
\end{aligned}
\label{eq:problem-space}
\end{equation}
where the SLM with parameters \(\theta\) has accuracy \(\operatorname{Acc}(f(q); \theta)\) on the mapped input, and \(\mathcal{S}\) is a set of fundamentally similar question pairs. The constraints encourage state compression and enforce a standardized structure within the problem space.

\begin{figure}[!t]
    \centering
    \includegraphics[width=3.2in]{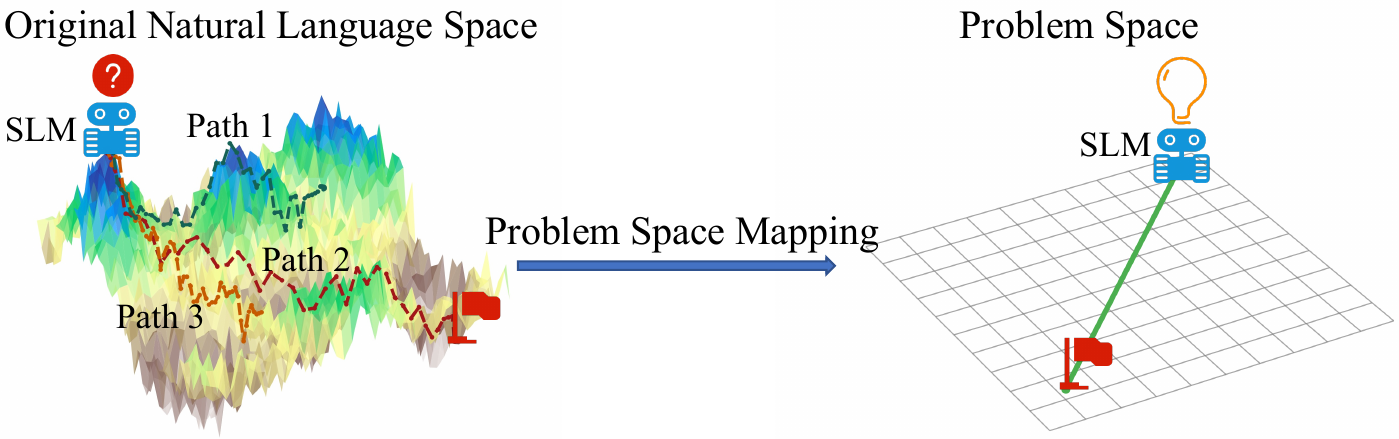}
    \caption{An illustration of our framework for decoupling understanding from reasoning. The original natural language space is complex and high-dimensional, making exploration difficult; mapping to a standardized, low-dimensional problem space compresses the state space and facilitates more efficient exploration.}
    \label{fig:framework}
\end{figure}

Based on this formulation, we propose a unified framework (Figure~\ref{fig:framework}) that leverages a dedicated problem space mapper to project natural language questions into a standardized representation. By clustering fundamentally similar problems, this mapping reduces the exploration space and improves both sample and exploration efficiency during SLM training. As the model advances within this space, its ability to solve more complex problems increases, gradually shifting the underlying problem distribution. To adapt, our framework adopts an iterative training paradigm that alternates between updating the problem space mapper and refining the reasoning model, enabling their co-evolution. Reducing the problem space dimensionality enhances exploration and speeds up convergence. Follow \cite{cui2025entropy}, we model the problem using a bandit setting and apply a simplified Upper Confidence Bound, showing that the regret bound decreases with problem space dimensionality through the following theorem.

\begin{theorem}
\label{theorem1}
Let $\mathcal{Q}$ be a finite set of natural language problems, viewed as distinct states $s$, and let $A$ denote the set of candidate responses. At each round $t \in \{1, \dots, T\}$, a SLM observes a problem $s_t \in \mathcal{Q}$, selects an action $a_t \in A$, and receives a reward $r(s_t, a_t)$. Suppose learning is performed via a state-wise Upper Confidence Bound (UCB) algorithm in a contextual bandit setting. Then, in the state-independent worst case, the total regret after $T$ rounds is bounded by
\[
R_T = O\left( \sqrt{|\mathcal{Q}| \cdot |A| \cdot T \cdot \ln T} \right),
\]
where $|.|$ is the number of element of the set.
\end{theorem}
The proof is provided in the Appendix A\ref{appendix:appendix_theorem1_proof}. While the UCB setting simplifies that of LLMs, Theorem~\ref{theorem1} offers valuable insight into how reduced problem space dimensionality improves exploration. Specifically, mapping from \( \mathcal{Q} \) to \( \mathcal{P} \) compresses the space by a ratio \( \alpha = |\mathcal{P}| / |\mathcal{Q}| < 1 \), tightening the regret bound by a factor of \( \sqrt{\alpha} \). This result supports our central motivation: leveraging standardized abstraction can make reasoning training more efficient for SLMs.

\section{Methods}
\begin{figure*}[htp]
    \centering
    \includegraphics[width=1.92\columnwidth]{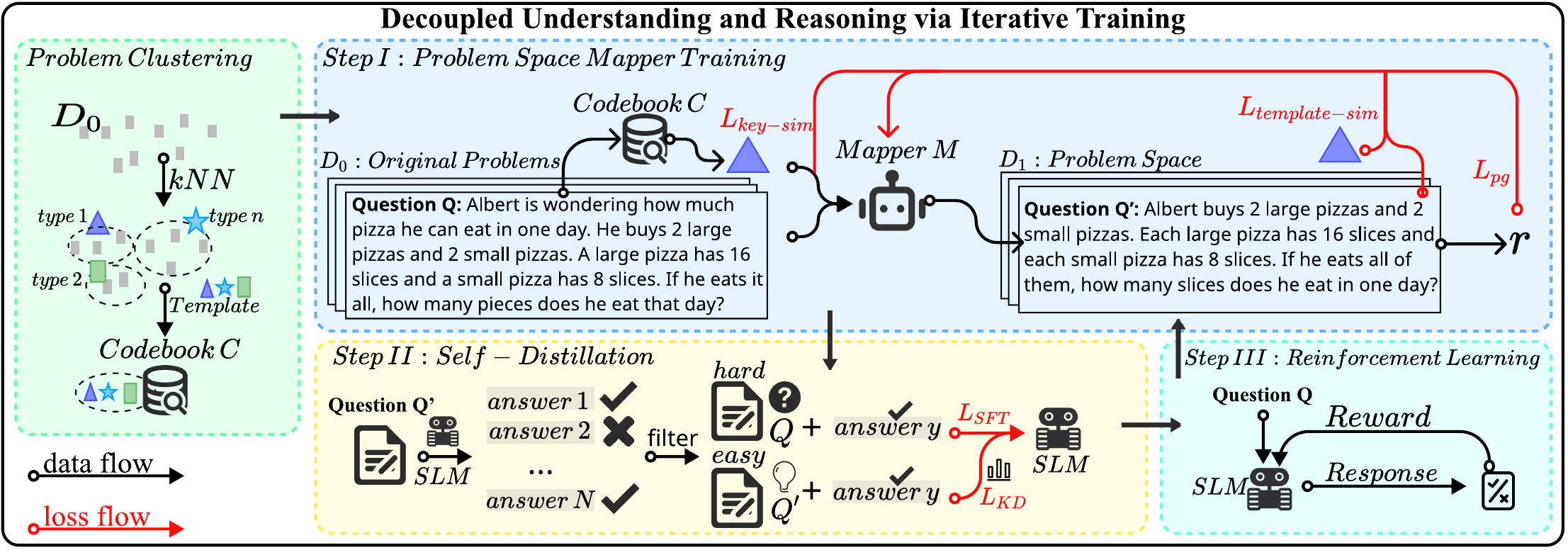} 
    \caption{Framework of the DURIT Method. After KNN-based clustering, DURIT (1) compresses problems via implicit mapping, (2) distills this into the SLM, and (3) optimizes it through reinforcement learning with alternating co-training.}
    \label{fig:durit_method}
\end{figure*}

We propose Decoupled Understanding from Reasoning via Iterative Training (DURIT), which designs to enhance the reasoning ability of SLMs by decoupling problem understanding from reasoning. As illustrated in Figure~\ref{fig:durit_method}, DURIT consists of three alternating steps: (1) \textbf{Problem Mapper Training:} a problem mapper $M$ is trained via RL, guided by implicit templates, to map original natural language problems into problem space. (2) \textbf{Self-Distillation:} The transformation capability is internalized into reason the SLM $R$ via self-distillation, enabling it to directly process complex problems without reliance on the external mapper $M$ at inference time. (3) \textbf{RL Training:} the SLM $R$ is further optimized using RL to improve its reasoning performance. The three steps are repeated iteratively, progressively strengthening the model's reasoning through alternating phases of understanding and reasoning. The complete pseudocode is provided in the Appendix C\ref{appendix:pseudocode}.

\subsection{Step I: Problem Space Mapper Training}
To decouple understanding from reasoning, a problem space mapper \( M \) is instantiated as an LLM that maps natural language questions into a standardized problem space. While explicit templates enforce standardization, they are labor-intensive and may impede comprehension by SLMs. To balance standardization and flexibility, an implicit template mechanism is proposed, using a codebook to softly guide the output style of $M$. The mapping aims to 1) improve SLMs' understanding and 2) reduce the complexity of the problem space. To facilitate this, we cluster the training data based on fundamental question similarity using k-Nearest Neighbors (kNN) over representations \( \mathbf{z}_i \) encoded from each question \( Q_i \), its description, and answer via model \( M \). As no ground-truth labels exist, we adopt GRPO~\cite{shao2024deepseekmathpushinglimitsmathematical} to optimize \( M \) based on the average correctness $r_{\text{acc}}$ of frozen SLM’s responses to mapped problem \( Q_i' \). To prevent \( M \) from solving the problem directly, we apply a cheating penalty $r_{\text{cheating}}$ if \( Q_i' \) includes solution-specific terms (e.g., keywords like “answer value”) not present in \( Q_i \). The total reward is:
\begin{equation}
r_i = r_{\text{acc}} + r_{\text{cheating}}.
\end{equation}

However, RL alone cannot sufficiently enforce standardization. To simplify the problem space, implicit templates conditioned on cluster labels \( t_i \) are introduced. Specifically, a codebook  $C$ of \( n \) implicit template tokens \( \{T_1, \ldots, T_n\} \) and corresponding query keys \( \{k_1, \ldots, k_n\} \) is constructed, with both \( \{T_i\} \) and \( \{k_i\} \) randomly initialized parameters and optimized by loss. During training, for each problem \( Q_i \), the template token \( T_{t_i} \) is selected and concatenated with the original input as \( x_i = [Q_i; T_{t_i}] \), guiding \( M \) to produce the mapped question \( Q'_i \). To encourage alignment between \( Q'_i \) and its assigned template, we define a template similarity loss based on the InfoNCE~\cite{He_2020_CVPR} objective:
\begin{equation}
\mathcal{L}_{\text{template-sim}} = - \log \frac{\exp\left( \frac{ \langle \mathbf{z}_i, \mathbf{T_{t_i}} \rangle }{\tau} \right)}{ \sum_{j=1}^n \exp\left( \frac{ \langle \mathbf{z}_i, \mathbf{T}_{j} \rangle }{\tau} \right)},
\end{equation}
where \( \langle \cdot, \cdot \rangle \) denotes the inner product, $\mathbf{z}_i$ is the normalized representation of the mapped problem $Q'_i$ and $\tau$ is a temperature hyperparameter. At inference, with \( t_i \) unavailable, the best-matching implicit template is selected via cosine similarity between the input question embedding \( \mathbf{q}_i \) (both $\mathbf{z}_i$ and $\mathbf{q}_i$ are approximated by averaging the word embeddings) and learned template query keys. A key similarity loss is introduced to facilitate key learning:
\begin{equation}
\mathcal{L}_{\text{key-sim}} = - \log \frac{\exp\left( \frac{ \langle \mathbf{q}_i, \mathbf{k_i} \rangle }{\tau} \right)}{ \sum_{j=1}^n \exp\left( \frac{ \langle \mathbf{q}_i, \mathbf{k}_{j} \rangle }{\tau} \right)},
\end{equation}
Gradients from \( \mathbf{k}_i \) are detached to prevent interference with the training of \( M \), and only the template keys are updated. The overall loss function jointly optimizes the mapping policy and template-based constraints:
\begin{equation}
\mathcal{L}_{\text{total}} = \mathcal{L}_{\text{pg}} + \alpha_1 \mathcal{L}_{\text{key-sim}} + \alpha_2 \mathcal{L}_{\text{template-sim}},
\end{equation}
where $\mathcal{L}_{\text{pg}}$ denotes the policy gradient loss from GRPO, and $\alpha_1$, $\alpha_2$ are hyperparameters balancing different losses.

\subsection{Step II: Self-Distillation Training}
After training the problem space mapper \( M \), its transformation capability is internalized into the SLM via self-distillation. Specifically, \( M \) transforms the original dataset \( \mathcal{D}_0 \) into a normalized form $\mathcal{D}_1 = \{ Q_i' = M(Q_i) \mid Q_i \in \mathcal{D}_0 \}$, where the mapped questions are designed to facilitate easier reasoning for the SLM. We then sample \( N \) responses using SLM on each \( Q_i' \in \mathcal{D}_1 \), and construct a filtered dataset: $\mathcal{D}_2 = \{(Q_i, Q_i', y_i) \mid Q_i \in \mathcal{D}_0,\ y_i = R(Q_i'),\ \texttt{answer}(y_i) = \texttt{True} \}$, where \( y_i \) denotes the model's response and \texttt{answer}\((y_i)\) evaluates its correctness. The core idea is to encourage SLM to replicate on \( Q_i \) the reasoning behavior it exhibits on \( Q_i' \). To achieve this, we treat \( (Q_i', y_i) \) as a teacher pair and \( (Q_i, y_i) \) as the corresponding student pair. The SLM is trained using a combined loss of supervised fine-tuning \(L_{\mathrm{SFT}}\) and KD \(L_{\mathrm{KD}}\):
\begin{equation}
p(x^k) = \frac{\exp(x^k / \tau)}{\sum_{j=1}^{|V|} \exp(x^j / \tau)},
\end{equation}
\begin{equation}
\begin{split}
\mathcal{L}_i &= \frac{1}{l}\sum_{k=1}^l \Big[
    (1-\lambda)\big(-\log p_s(x_i^k)\big)\\
    &\qquad\qquad +\ \lambda\,\mathrm{KL}\!\big(p_t(x_i^k)\parallel p_s(x_i^k)\big)
\Big].
\end{split}
\end{equation}
where \( l \) is the sequence length, \( x_i^k \) the \( k \)-th token of \( y_i \), and \( p_s \), \( p_t \) the student’s and teacher’s softmax outputs for prefix inputs \( Q_i \) and \( Q'_i \), respectively. The parameter \( \lambda \) balances the losses. This setup allows the student to internalize \( M \) without accessing it at inference.

\subsection{Step III: Reinforcement Learning Training}
After distilling the transformation capability into the SLM, the model is further trained via RL to explore and reason directly in the original problem space, leveraging its internalized understanding. Specifically, we fine-tune the SLM on the original training dataset $\mathcal{D}_0$ using the GRPO algorithm, with answer correctness serving as the reward signal. As the reasoning model improves, its ability to interpret and generalize evolves, potentially altering the optimal structure of the problem space. To accommodate this, the problem space mapper $M$ and the reasoning model $R$ are trained iteratively, enabling continual refinement of the problem space and progressive enhancement of reasoning capabilities.

\section{Experiments}
\subsection{Datasets}
All experiments train models solely on GSM8K~\cite{cobbe2021trainingverifierssolvemath}. Evaluation considers both in-domain (IND) and out-of-domain (OOD) settings: GSM8K-Platinum~\cite{vendrow2025largelanguagemodelbenchmarks} for IND, and MAWPS~\cite{koncel-kedziorski-etal-2016-mawps}, SVAMP~\cite{patel2021nlpmodelsreallyable}, MATH500~\cite{hendrycks2021measuringmathematicalproblemsolving}, and GAOKAO~\cite{zhang2024evaluatingperformancelargelanguage} for OOD mathematical reasoning. Broader reasoning is evaluated on LogiQA~\cite{liu2020logiqachallengedatasetmachine}. This setup enables systematic analysis of DURIT’s impact on SLM reasoning across diverse domains.

\subsection{Baselines and Metric}
We compare DURIT against representative baselines in four categories: (1) CoT Distillation: including Std-CoT~\cite{magister-etal-2023-teaching} and STaR~\cite{zelikman2024star}, where $N$ CoT responses per question are sampled and correct ones are filtered for fine-tuning; (2) Prompt Optimization: PRewrite~\cite{kong-etal-2024-prewrite} using RL to optimize prompts; (3) RL-Based Methods: GRPO~\cite{shao2024deepseekmathpushinglimitsmathematical}; and (4) Knowledge Distillation: Vanilla-KD~\cite{NEURIPS2024_48229913}, which requires online teacher LM inference. In our setup, the mapper model serves as the teacher. Following prior work~\cite{Sheng_Li_Zeng_2025}, answer accuracy is the primary metric. Further baseline details are available in the Appendix B.3\ref{appendix:baseline_details}.

\subsection{Implementations}
To evaluate the generalization capability of DURIT, we test different base models, including recent strong instruction-following and reasoning-oriented models such as Qwen2.5-0.5B-Instruct \cite{yang2025qwen3} and Llama3.2-1B-Instruct \cite{grattafiori2024llama}. For the mapper model, we use Qwen2.5-3B-Instruct for the Qwen family, and Llama3.2-3B-Instruct for the Llama family to ensure architectural homogeneity. The codebook contains 32 implicit templates, with loss coefficients \( \alpha_1 = 1\mathrm{e}{-3} \) and \( \alpha_2 = 1\mathrm{e}{-2} \). Training is conducted in three steps: Step I runs for 1 epoch, Step II for 5 epochs, and Step III for 3 epochs. All experiments are carried out on 2 A100 GPUs with 40GB memory. For inference, we employ greedy decoding without vLLM \cite{kwon2023efficient} acceleration. Additional implementation details, as well as more experimental results, parameter analyses, and training time comparisons, can be found in Appendix E~\ref{appendix:more_exp_results}.

\subsection{Main Results}
As shown in Table~\ref{tab:main-results}, DURIT outperforms all baselines on both IND and OOD benchmarks. Remarkably, even when trained solely on the GSM8K dataset, DURIT consistently delivers substantial performance gains on all datasets. With just a single iteration, it achieves average accuracy improvements of 2.06\% and 2.35\% over the strongest baseline methods on Qwen2.5-0.5B-Instruct and Llama3.2-1B-Instruct, respectively. Importantly, DURIT achieves these gains without relying on external large models for CoT supervision. Instead, it fully exploits the model’s own reasoning abilities to explore, adapt, and transfer prior knowledge. Remarkably, DURIT even outperforms distillation-based methods that depend on stronger teacher models such as DeepSeek-R1. As it operates entirely within the model itself, DURIT avoids additional API costs and infrastructure overhead, offering broad applicability and high cost-efficiency. DURIT's reasoning ability is further enhanced through a second iteration of training: even when continuing to use the GSM8K dataset, it yields an average accuracy gain of 0.36\% on Qwen2.5-0.5B-Instruct and 0.69\% on Llama3.2-1B-Instruct. Greater improvements are observed when using different datasets in the second iteration (see later Section~\ref{sec:iter_diff_data}), demonstrating DURIT’s strong generalization across domains and its effectiveness in reducing the cognitive load of reasoning acquisition.

\begin{table*}[t]
\label{table:main_results}
\centering
\small
\setlength{\tabcolsep}{4.5pt}
\renewcommand{\arraystretch}{1.2}
\begin{tabular}{l|c|ccccc|c}
\toprule
\multirow{2}{*}{\textbf{Methods}} & \textbf{In-Domain} & \multicolumn{5}{c|}{\textbf{Out-of-Domain}} & \multirow{2}{*}{\textbf{Average}} \\
& gsm8k-platinum & MAWPS & SVAMP & MATH500 & GAOKAO & LogiQA &  \\
\midrule
\multicolumn{8}{l}{\textbf{\# Qwen2.5-0.5B-Instruct based}} \\
\midrule
Base \cite{yang2025qwen3} & 45.74 & 54.23 & 54.67 & 27.80 & 18.55 & 14.44 & 35.91 \\
CoT-Dis \cite{magister-etal-2023-teaching} & 44.67 & 55.77 & 58.33 & 18.80 & 12.90 & \textbf{30.41} & 36.81 \\
STaR \cite{zelikman2024star} & 51.86 & 57.88 & 61.67 & 29.60 & 18.55 & 23.50 & 40.51 \\
GRPO \cite{shao2024deepseekmathpushinglimitsmathematical} & 51.03 & 58.08 & 61.00 & 27.40 & 21.77 & 22.73 & 40.34 \\
PRewrite \cite{kong-etal-2024-prewrite} & 47.23 & 56.73 & 57.00 & 29.80 & 19.35 & 23.96 & 39.01 \\
Vanilla-KD \cite{NEURIPS2024_48229913} & 49.30 & 57.69 & 61.67 & 30.4 & \textbf{23.39} & 20.74 & 40.53 \\
DURIT (ours, iter=1) & \textbf{53.68} & \underline{60.19} & \underline{62.67} & \underline{31.00} & \textbf{23.39} & 24.58 & \underline{42.59} \\
DURIT (ours, iter=2) & \underline{53.10} & \textbf{60.38} & \textbf{63.00} & \textbf{32.80} & \underline{22.58} & \underline{25.81} & \textbf{42.95} \\
\midrule
\multicolumn{8}{l}{\textbf{\# Llama3.2-1B-Instruct based}} \\
\midrule
Base \cite{grattafiori2024llama} & 30.52 & 5.77 & 20.67 & 22.60 & 12.10 & 1.54 & 15.53 \\
CoT-Dis \cite{magister-etal-2023-teaching} & 48.06 & 56.92 & 57.67 & 24.60 & 12.90 & \textbf{21.81} & 36.99 \\
STaR \cite{zelikman2024star} & 36.31 & 52.50 & 54.33 & 20.00 & \textbf{16.94} & 8.45 & 31.42 \\
GRPO \cite{shao2024deepseekmathpushinglimitsmathematical} & 48.39 & 59.23 & 57.67 & \underline{26.40} & \underline{16.13} & 4.45 & 35.38 \\
PRewrite \cite{kong-etal-2024-prewrite} & 35.81 & 41.34 & 46.00 & 18.80 & 12.10 & 3.53 & 26.26 \\
Vanilla-KD \cite{NEURIPS2024_48229913} & 42.35 & \textbf{64.23} & 62.67 & 22.40 & \underline{16.13} & 7.99 & 35.96 \\
DURIT (ours, iter=1) & \underline{50.37} & 59.62 & \underline{64.33} & 26.00 & 14.52 & \underline{21.20} & \underline{39.34} \\
DURIT (ours, iter=2) & \textbf{52.36} & \underline{62.31} & \textbf{66.00} & \textbf{27.60} & 12.10 & 19.82 & \textbf{40.03} \\
\bottomrule
\end{tabular}
\caption{Performance (\%) of Qwen2.5-0.5B-Instruct and Llama3.2-1B-Instruct models across six representative benchmarks under various methods. The \textbf{bold} and \underline{underline} indicate the best and second-best results, respectively.}
\label{tab:main-results}
\end{table*}

\subsection{Reasoning Robustness Evaluation}
By projecting natural language questions into a more intrinsic and low-dimensional problem space, DURIT focuses on the essential semantics of the problem. This abstraction reduces variation from surface-level expressions and suppresses spurious or irrelevant cues, thereby enhancing the robustness of reasoning. To verify this claim, we evaluate on the GSM-Symbolic benchmark \cite{mirzadeh2024gsmsymbolicunderstandinglimitationsmathematical} using Qwen2.5-0.5B-Instruct and LLaMA3.2-1B-Instruct. As the original dataset contains only 100 examples and exhibits high variance, we follow \cite{gao2025abstralaugmentingllmsreasoning, liu2025cogmathassessingllmsauthentic} and adopt the relative drop in average accuracy as a robustness metric. Results are reported in Table~\ref{tab:robust-results}. DURIT attains an almost minimal relative drop in accuracy among all methods, indicating that its reasoning gains are accompanied by notably enhanced robustness. Results for additional model scales are provided in the Appendix E.2\ref{appendix:more_model_eval}.

\begin{table}[ht]
\centering
\small
\begin{tabular}{l|ccc|ccc}
\toprule
\multirow{2}{*}{Method} & \multicolumn{3}{c|}{Qwen-0.5B} & \multicolumn{3}{c}{Llama-1B} \\
\cmidrule{2-7}
 & Orig & Symb & $\Delta\%$ & Orig & Symb & $\Delta\%$ \\
\midrule
Base & 46.0 & 41.6 & \textbf{-9.6}  & 21.0 & 16.0 & -23.7 \\
CoT-Dis & 47.0 & 40.6 & -13.7   & 51.0 & 38.3 & -24.9 \\
STaR & 51.0 & 41.0 & -19.7 & 33.0 & 27.1 & \underline{-17.8} \\
GRPO & 50.0 & 42.9 & -14.3 & 44.0 & 35.8 & -18.6 \\
PRewrite & 48.0 & 42.0 & -12.0     & 39.0 & 21.9 & -43.8 \\
Vanilla-KD & 51.0 & 42.2 & -17.2 & 42.0 & 33.4 & -20.5 \\
DURIT & 48.0 & 42.6 & \underline{-11.3} & 44.0 & 40.8 & \textbf{-7.2} \\
\bottomrule
\end{tabular}
\caption{Comparison of methods on Qwen2.5-0.5B-Instruct and Llama3.2-1B-Instruct. DURIT is trained with a single iteration. Orig: original test set; Symb: gsm-symbolic; $\Delta\%$: relative drop from Orig to Symb. \textbf{Bold} and \underline{underline} indicate best and second-best results.}
\label{tab:robust-results}
\end{table}

\subsection{Performance Across Different Iterative Training Data}
\label{sec:iter_diff_data}
\begin{figure*}[!t]
    \centering
    \begin{subfigure}[t]{0.33\textwidth}
        \centering
        \includegraphics[height=3.3cm, width=\linewidth]{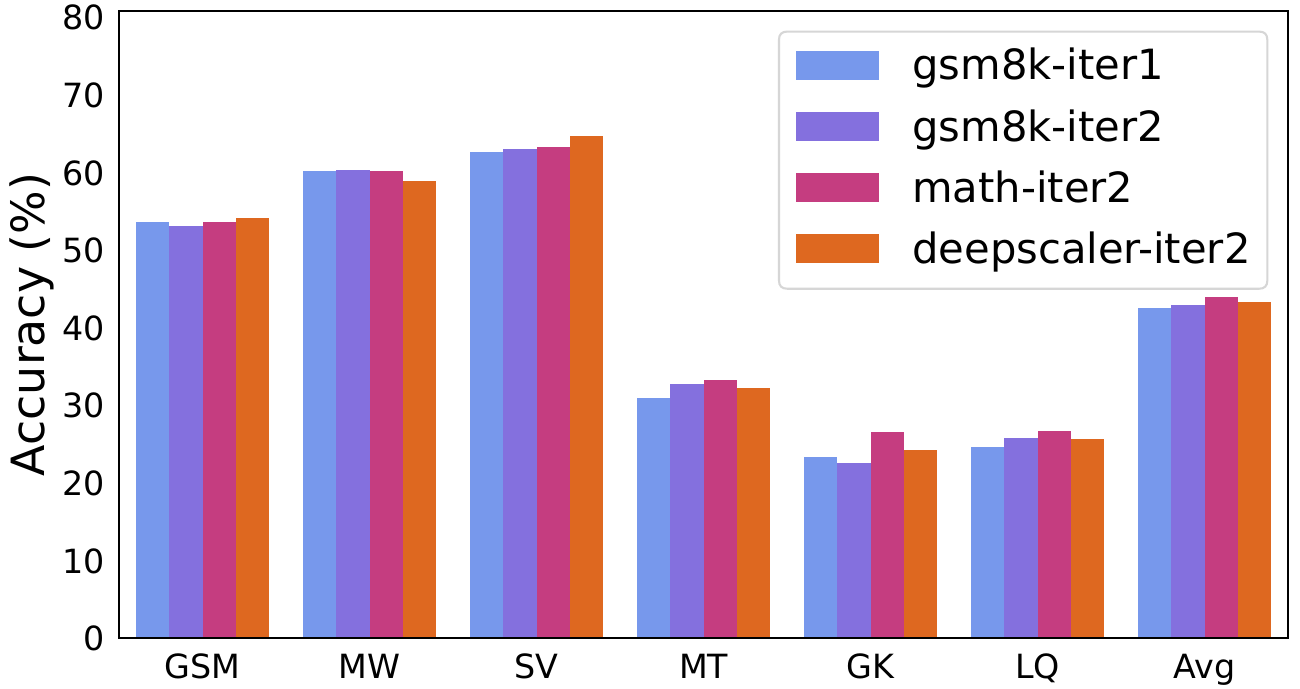}
        \caption{Impact of different iterative training data.}
        \label{fig:diff_iter_data}
    \end{subfigure}
    \hfill
    \begin{subfigure}[t]{0.33\textwidth}
        \centering
        \includegraphics[width=\linewidth]{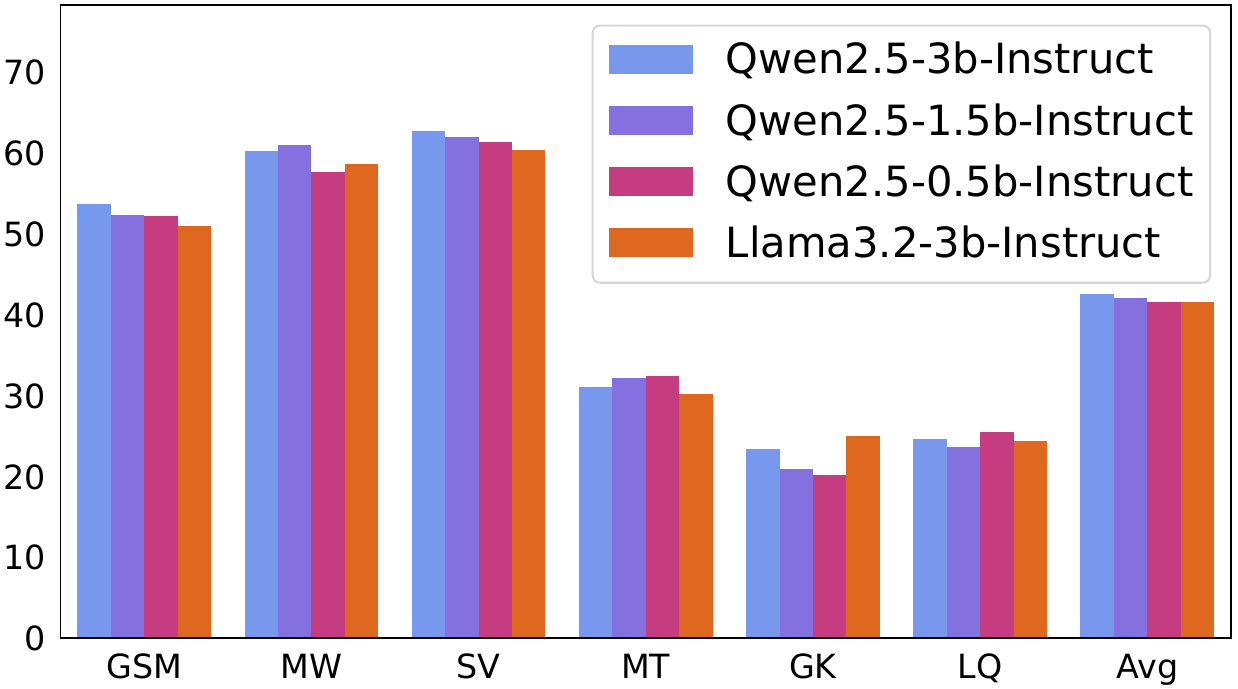}
        \caption{Impact of different mapper models.}
        \label{fig:diff_mapper}
    \end{subfigure}
    \hfill
    \begin{subfigure}[t]{0.33\textwidth}
        \centering
        \includegraphics[width=\linewidth]{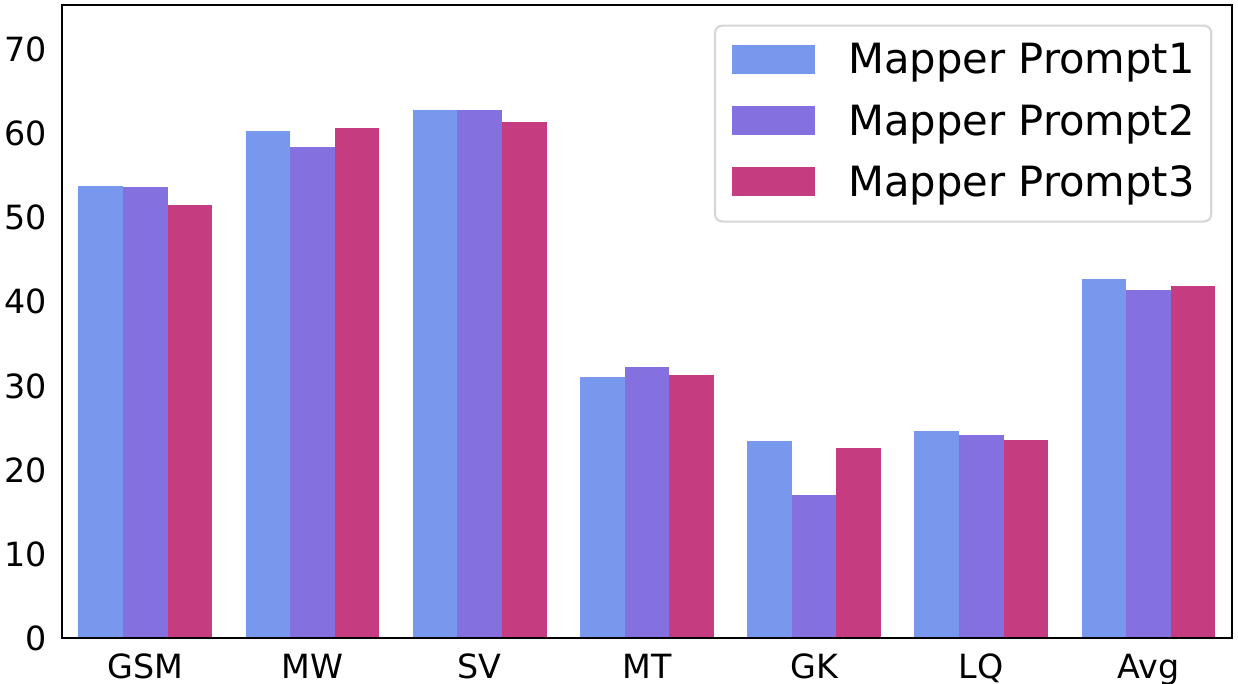}
        \caption{Impact of different mapper prompt.}
        \label{fig:diff_mapper_prompt}
    \end{subfigure}
    \caption{DURIT performance across six benchmarks with varying training data, mapper models, and prompts: GSM (GSM8K), MW (MAWPS), SV (SVAMP), MT (MATH500), GK (GAOKAO), LQ (LogiQA), and average (Avg).}
    \label{fig:diff_combined}
\end{figure*}
To evaluate the impact of iterative training datasets on DURIT, we conducted a second iteration using GSM8K, MATH, and a filtered DeepScaleR~\cite{deepscaler2025}, following the first iteration on GSM8K. As Figure~\ref{fig:diff_iter_data} shows, improvements are more pronounced when the second-iteration data differs from the first. This demonstrates DURIT’s ability to decouple understanding from reasoning, effectively leveraging complementary data. Additionally, training with more diverse datasets consistently enhances overall performance and reasoning capabilities. Dataset details are provided in the Appendix B.2\ref{appendix:data_iter_exp}.

\subsection{Performance Across Different Mapper Models}
To evaluate the impact of different mappers on DURIT, we fix the reasoning SLM as Qwen2.5-0.5B-Instruct and perform one iteration of DURIT updates with various mappers (Qwen2.5-3B/1.5B/0.5B-Instruct and Llama3.2-3B-Instruct) to assess both model scale and family effects. Given the relatively weak instruction-following of Qwen2.5-0.5B-Instruct, we warm-start it with 200 mapper data from Qwen2.5-3B-Instruct to improve initial alignment. Results (Figure~\ref{fig:diff_mapper}) show that mappers within the same family generally outperform others, and performance slightly improves with larger model size. Overall differences are marginal, demonstrating DURIT’s robustness to mapper choice: even with a lightweight mapper like Qwen2.5-0.5B-Instruct, strong performance is achieved without relying on external larger models.

\subsection{Performance Across Different Mapper Prompts}
To assess the impact of mapper prompt design on DURIT, we test three prompt formulations (see Appendix B.3\ref{appendix:baseline_details}) using Qwen2.5-3B-Instruct as the mapper and Qwen2.5-0.5B-Instruct as the reasoning SLM, training each configuration for one iteration. As Figure~\ref{fig:diff_mapper_prompt} shows, performance varies slightly: more explicit, standardized prompts generally perform better. All variants achieve strong performance, indicating DURIT’s robustness to mapper prompt variations.

\subsection{Ablation Studies}
We conduct ablation studies using the Qwen2.5-0.5B-Instruct model to evaluate the contribution of each component in DURIT. For Step~I, we assess the impact of removing the implicit template constraint component (\texttt{w/o tem}), while keeping the subsequent procedures in Step~II and Step~III of DURIT unchanged. For Step II, we examine the role of self-distillation by removing it and retaining only the SFT loss (\texttt{w/o sd}) and the necessity of SFT by removing sft loss(\texttt{w/o sft}). For Step III, we investigate the effect of removing GRPO training (\texttt{w/o grpo}). As shown in Table~\ref{tab:ablation}, ablating any single component leads to performance degradation. \texttt{w/o tem} disrupts the standardization of the problem space, resulting in less compact representations and lower exploration efficiency. \texttt{w/o sd} has minimal impact on in-domain performance but substantially impairs out-of-domain generalization, underscoring the role of self-distillation in reducing the comprehension burden and enhancing robustness. \texttt{w/o sft} may impose excessive disruption on the model’s inherent reasoning mechanisms and simultaneously expose it to biased or incorrect reasoning patterns in the mapped questions, potentially resulting in further performance degradation. Finally, \texttt{w/o grpo} consistently reduces accuracy, confirming the necessity of RL to strengthen reasoning after self-distillation.
\begin{table}[t]
\centering
\small  
\setlength{\tabcolsep}{3pt}  
\renewcommand{\arraystretch}{1.0}  
\begin{tabular}{l|c|ccccc|c}
\toprule
\textbf{Variant} & \textbf{GSM} & \textbf{MW} & \textbf{SV} & \textbf{MT} & \textbf{GK} & \textbf{LQ} & \textbf{Avg.} \\
\midrule
DURIT & \textbf{53.68} & 60.19 & \textbf{62.67} & 31.00 & \textbf{23.39} & 24.58 & \textbf{42.59} \\
\texttt{w/o tem} & 53.02 & 59.04 & 60.00 & \textbf{31.40} & 20.97 & 23.04 & 41.25 \\
\texttt{w/o sd} & 53.52 & \textbf{60.58} & 58.33 & 30.80 & 21.77 & \textbf{24.88} & 41.65 \\
\texttt{w/o sft} & 51.20 & 57.31 & 61.67 & 28.60 & 19.35 & 22.73 & 40.14 \\
\texttt{w/o grpo} & 49.30 & 57.69 & 61.67 & 30.40 & \textbf{23.39} & 20.74 & 40.53 \\
\bottomrule
\end{tabular}
\caption{Ablation study of Qwen2.5-0.5B-Instruct on six benchmarks with a single DURIT iteration.}
\label{tab:ablation}
\end{table}

\subsection{Towards Understanding the Effectiveness of Problem Space Mapping}
\begin{figure}[t]
  \centering
  \begin{minipage}[t]{0.48\linewidth}
    \centering
    \includegraphics[width=\linewidth]{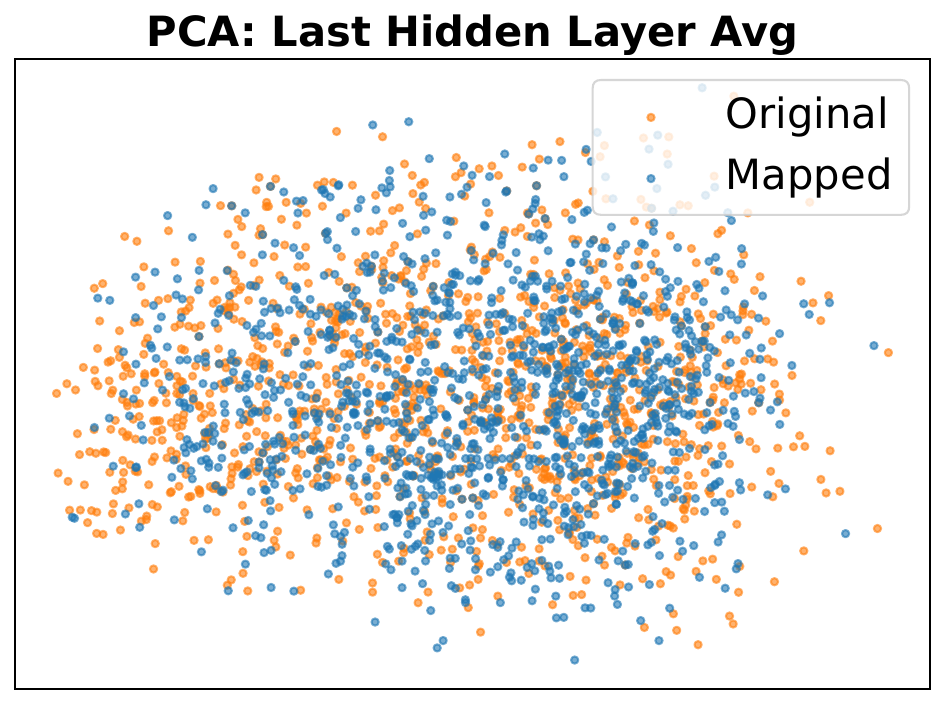}
    \caption*{(a) PRewrite PCA}
  \end{minipage}
  \hfill
  \begin{minipage}[t]{0.48\linewidth}
    \centering
    \includegraphics[width=\linewidth]{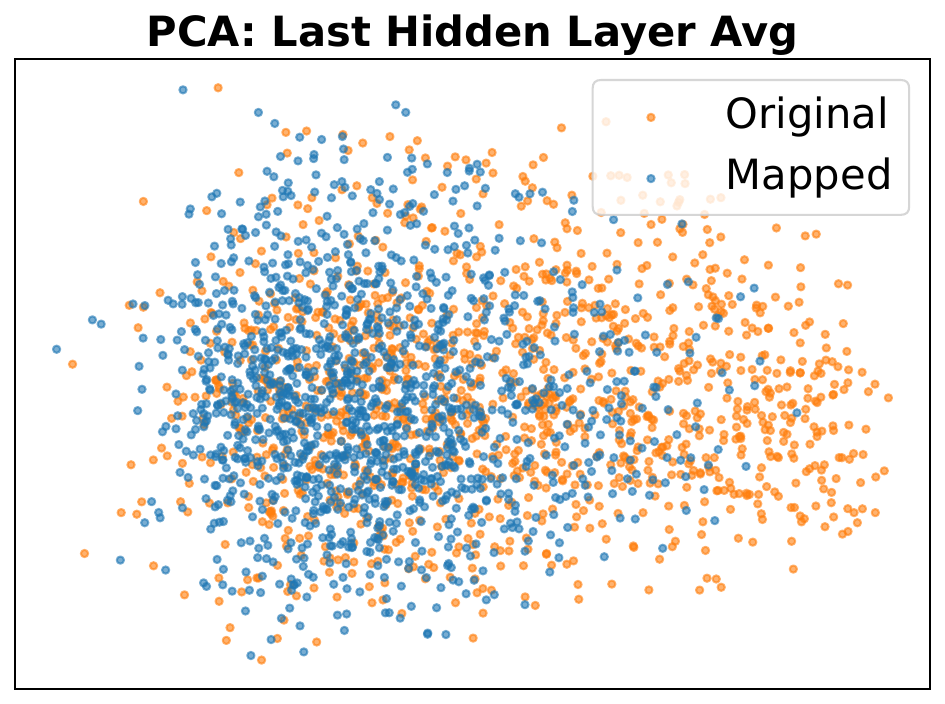}
    \caption*{(b) DURIT PCA}
  \end{minipage}
  \caption{PCA Visualizations of Hidden Representations from Different Methods Using Qwen2.5-0.5B-Instruct.}
  \label{fig:pca-comparison}
\end{figure}

To visualize how mapped questions are represented within the SLM, we analyze the final hidden layer representations of Qwen2.5-0.5B-Instruct on GSM8K-Platinum, using both the original inputs and their mapped versions produced by PRewrite and DURIT. We quantify the local compactness of these representations using the average k-nearest neighbor distance, as reported in Table~\ref{tab:knn-distance}. Additionally, we apply Principal Component Analysis (PCA) to project the high-dimensional hidden states into 2D for visualization, as shown in Figure~\ref{fig:pca-comparison}. Compared to the original and PRewrite-mapped inputs, DURIT-mapped inputs yield significantly more compact clusters in the embedding space. This suggests that DURIT mapping helps remove redundant or irrelevant linguistic variability, effectively reducing the dimensionality of the problem space. As a result, the model may better capture the underlying essence of the problems, potentially leading to more efficient learning. 

\begin{table}[t]
  \centering
  \begin{tabular}{lccc}
    \toprule
    Input & Original & PRewrite & DURIT \\
    \midrule
    5NN Distance & 75.16 & 73.68 & \textbf{68.59} \\
    \bottomrule
  \end{tabular}
  \caption{Average 5-NN distance of Qwen2.5-0.5B-Instruct final hidden states across different input questions. Lower values indicate tighter local clusters.}
  \label{tab:knn-distance}
\end{table}

\section{Conclusion}
In this work, we propose a general problem space mapping framework, upon which we instantiate a concrete algorithm DURIT. DURIT consists of three key steps: (1) a problem space mapper trained via reinforcement learning with implicit template guidance, (2) self-distillation to internalize the mapping capability into a SLM, and (3) reasoning optimization of SLM within the reduced problem space. By alternating the training of the mapper and the SLM, DURIT enables iterative improvements in both reasoning capability and robustness. Empirical results demonstrate that DURIT consistently outperforms fine-tuned baselines, achieving substantial improvements in both in-domain and out-of-domain reasoning tasks, as well as enhanced robustness.

\section{Acknowledgments}
This work was supported by the National Science and Technology Major Project (Grant No. 2022ZD0117402), the National Natural Science Foundation of China (Grant No. 62441617), and the Beijing Advanced Innovation Center for Future Blockchain and Privacy Computing.

\bibliography{aaai2026}

\appendix
\setcounter{theorem}{0}
\section{Appendix A: Proof of Theorem 1}
\label{appendix:appendix_theorem1_proof}
\begin{theorem}
\label{appendix:theorem1}
Let $\mathcal{Q}$ be a finite set of natural language problems, viewed as distinct states $s$, and let $A$ denote the set of candidate responses. At each round $t \in \{1, \dots, T\}$, a SLM observes a problem $s_t \in \mathcal{Q}$, selects an action $a_t \in A$, and receives a reward $r(s_t, a_t)$. Suppose learning is performed via a state-wise Upper Confidence Bound (UCB) algorithm in a contextual bandit setting. Then, in the state-independent worst case, the total regret after $T$ rounds is bounded by
\[
R_T = O\left( \sqrt{|\mathcal{Q}| \cdot |A| \cdot T \cdot \ln T} \right),
\]
where $|.|$ is the number of element of the set.
\end{theorem}

\begin{proof}
For each natural language problem \( x \in \mathcal{Q} \), a SLM \( m \) with a token limit \( T \) and vocabulary \( V \) has up to \( V^T \) possible responses, forming a finite bandit problem. Thus, we treat each problem \( x \) as a distinct state \( s \) and define the candidate response set as the action space \( A \). This formulation is also aligned with commonly adopted outcome reward theoretical analyses~\cite{cui2025entropy} and algorithms~\cite{shao2024deepseekmathpushinglimitsmathematical,yu2025dapo}. At each round $t \in \{1, \dots, T\}$, the agent observes a state $s_t \in \mathcal{Q}$, selects an action $a_t \in A$, and receives a reward $r_t = r(s_t, a_t) \in [0, 1]$ with expected value $\mu(s_t, a_t)$. Let $\mu^*(s) = \max_{a \in A} \mu(s, a)$ be the optimal expected reward for state $s$. The total regret is defined as:
\[
R_T = \sum_{t=1}^T \left[ \mu^*(s_t) - \mu(s_t, a_t) \right].
\]

\textbf{Step 1: Decompose regret by state-action pairs.}  
Let $\Delta(s, a) = \mu^*(s) - \mu(s, a)$ denote the suboptimality gap for action $a$ in state $s$. Let $N_T(s, a)$ be the number of times action $a$ is selected in state $s$ up to round $T$. Then, the total regret can be expressed as:
\[
R_T = \sum_{s \in \mathcal{Q}} \sum_{a \neq a^*(s)} \Delta(s, a) \cdot \mathbb{E}[N_T(s, a)],
\]
where $a^*(s) = \arg\max_{a \in A} \mu(s, a)$.

\textbf{Step 2: Bound regret for a single state.}  
Fix $s \in \mathcal{Q}$, and let $T_s = \sum_{a \in A} N_T(s, a)$ be the number of times state $s$ occurs. When in state $s$, the agent faces a standard multi-armed bandit problem with $|A|$ arms. By the UCB algorithm, in the state-independent worst case, the regret for state $s$ is bounded by \cite{auer2002finite, lattimore2020bandit}:
\[
R_s = \sum_{a \neq a^*(s)} \Delta(s, a) \cdot \mathbb{E}[N_T(s, a)] = O\left( \sqrt{|A| \cdot T_s \cdot \ln T_s} \right).
\]
Since $T_s \leq T$, we have $\ln T_s \leq \ln T$, and thus:
\[
R_s = O\left( \sqrt{|A| \cdot T_s \cdot \ln T} \right).
\]

\textbf{Step 3: Sum over all states.}  
The total regret is the sum of the regrets over all states:
\[
R_T = \sum_{s \in \mathcal{Q}} R_s = \sum_{s \in \mathcal{Q}} O\left( \sqrt{|A| \cdot T_s \cdot \ln T} \right).
\]
Let $C_1 > 0$ be a constant such that $R_s \leq C_1 \sqrt{|A| \cdot T_s \cdot \ln T}$ for all $s \in \mathcal{Q}$. Then:
\[
R_T \leq C_1 \sqrt{|A| \cdot \ln T} \cdot \sum_{s \in \mathcal{Q}} \sqrt{T_s}.
\]

\textbf{Step 4: Apply the Cauchy-Schwarz inequality.}  
To bound the sum $\sum_{s \in \mathcal{Q}} \sqrt{T_s}$, we use the Cauchy-Schwarz inequality:
\[
\sum_{s \in \mathcal{Q}} \sqrt{T_s} \leq \sqrt{ |\mathcal{Q}| \cdot \sum_{s \in \mathcal{Q}} T_s }.
\]
Since $\sum_{s \in \mathcal{Q}} T_s = T$, this simplifies to:
\[
\sum_{s \in \mathcal{Q}} \sqrt{T_s} \leq \sqrt{|\mathcal{Q}| \cdot T}.
\]

\textbf{Step 5: Final bound.}  
Substituting back into the expression for $R_T$, we obtain:
\[
R_T \leq C_1 \sqrt{|A| \cdot \ln T} \cdot \sqrt{|\mathcal{Q}| \cdot T} = C_1 \sqrt{|\mathcal{Q}| \cdot |A| \cdot T \cdot \ln T}.
\]
Therefore, the total regret is bounded by:
\[
R_T = O\left( \sqrt{|\mathcal{Q}| \cdot |A| \cdot T \cdot \ln T} \right).
\]
\end{proof}

\section{Appendix B: Experimantal Settings}
\label{appendix:train_details}

\subsection{Appendix B.1: Datasets for DURIT Main Experiments}
\label{appendix:dataset_details}
\begin{figure*}[htp]
    \centering
    \includegraphics[width=2\columnwidth]{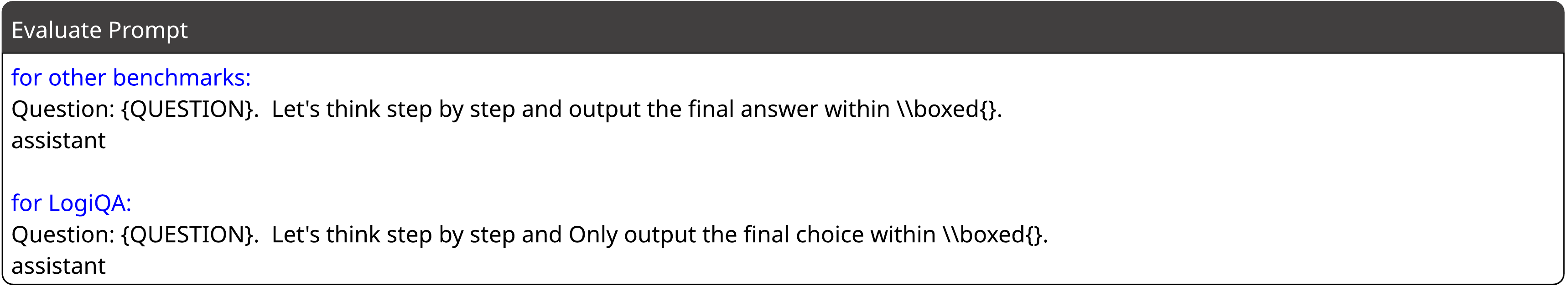} 
    \caption{Prompt used for evaluating all models.}
    \label{fig:eval_prompt}
\end{figure*}

We summarize the number of samples in the training and evaluation datasets in Table~\ref{tab:eval_datasets}. The GSM8K \cite{cobbe2021trainingverifierssolvemath} dataset is used as the primary training set. GSM8K-1 to GSM8K-3 refers to the filtered subset obtained in Step II of DURIT, while GSM8K-4 contains CoT examples generated by DeepSeek-R1 \cite{shao2024deepseekmathpushinglimitsmathematical}. GSM8K-5 through GSM8K-7 correspond to CoT data generated by different base models for STaR. All other datasets are used exclusively for evaluation with prompt in Figure~\ref{fig:eval_prompt}.
\begin{table}[t]
\centering
\small
\setlength{\tabcolsep}{6pt}
\renewcommand{\arraystretch}{1.1}
\begin{tabular}{lll}
\toprule
\textbf{Dataset} & \textbf{Description
} & \textbf{\# Train / Test} \\
\midrule
GSM8K            & Math Reasoning     & 7473 / - \\
GSM8K-1           & DURIT(Qwen0.5b-Step II)     & 6230 / - \\
GSM8K-2           & DURIT(Qwen1.5b-Step II)     & 7030 / - \\
GSM8K-3           & DURIT(Llama1b-Step II)     & 5770 / - \\
GSM8K-4           & DeepSeek-R1-CoT     & 7088 / - \\
GSM8K-5           & Qwen-0.5B(Self-CoT)     & 5714 / - \\
GSM8K-6         & Qwen-1.5B(Self-CoT)     & 7208 / - \\
GSM8K-7         & Llama-1B(Self-CoT)     & 6109 / - \\
GSM8K-platinum   & Math Reasoning     & - / 1209 \\
GSM-symbolic   & Math Reasoning     & - / 5000 \\
MAWPS            & Math Reasoning     & - / 520 \\
SVAMP            & Math Reasoning     & - / 300 \\
MATH500          & Math Reasoning     & - / 500 \\
GAOKAO           & Math Reasoning     & - / 125 \\
LogiQA           & Logical Reasoning  & - / 651 \\
\bottomrule
\end{tabular}
\caption{Statistics of evaluation datasets. GSM8K is used for training; all others are used for testing.}
\label{tab:eval_datasets}
\end{table}

\subsection{Appendix B.2: Datasets for Iterative Training Experiments}
\label{appendix:data_iter_exp}
The datasets used in the iterative training experiments are described as follows. In the second iteration, we used the full 7,473 samples from the GSM8K dataset and the complete 7,500 samples from the MATH dataset \cite{hendrycks2021measuringmathematicalproblemsolving}. For DeepScaleR \cite{deepscaler2025}, to control the difficulty level, we first generated 8 responses per instance using a temperature of 0.7 and accelerated decoding with vLLM \cite{kwon2023efficient}. We then discarded samples exhibiting extreme difficulty—specifically, those with an average accuracy of 1.0 or 0.0 across the sampled responses. From the remaining examples, we randomly selected 7,500 samples for training, ensuring that all three experimental settings used datasets of equal size.

\subsection{Appendix B.3: Training and Comparison Details}
\label{appendix:baseline_details}
\begin{figure*}[htp]
    \centering
    \includegraphics[width=2\columnwidth]{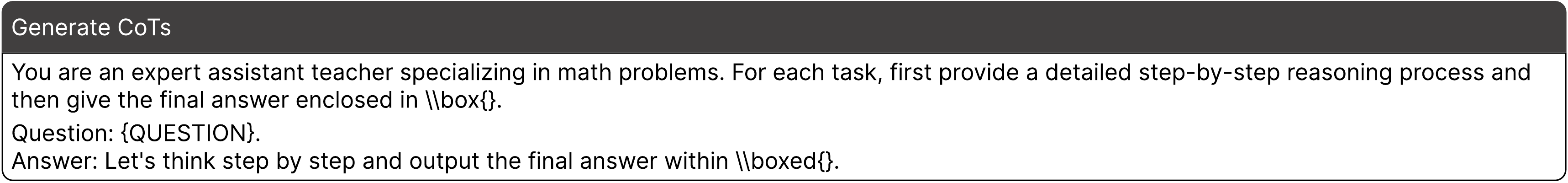} 
    \caption{Prompt used to guide DeepSeek-R1-0528 for generating CoT reasoning traces.}
    \label{fig:cot_prompt}
\end{figure*}

\begin{figure*}[htp]
    \centering
    \includegraphics[width=2\columnwidth]{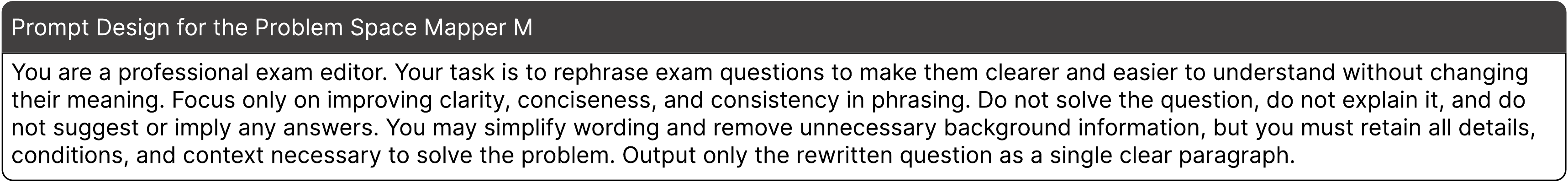} 
    \caption{Prompt of the Problem Space Mapper $M$.}
    \label{fig:mapper_prompt}
\end{figure*}

\begin{figure*}[htp]
    \centering
    \includegraphics[width=2\columnwidth]{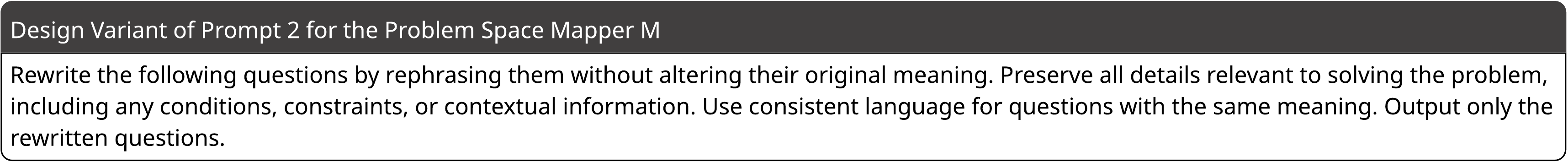} 
    \caption{Exploratory Prompt 2 for Problem Space Mapper $M$.}
    \label{fig:mapper_prompt2}
\end{figure*}

\begin{figure*}[htp]
    \centering
    \includegraphics[width=2\columnwidth]{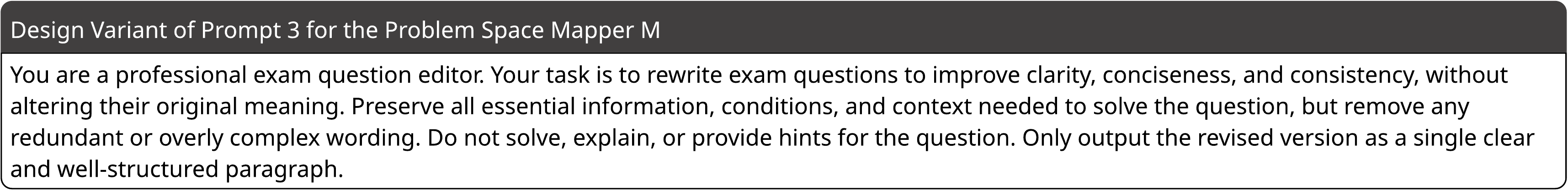} 
    \caption{Exploratory Prompt 3 for Problem Space Mapper $M$.}
    \label{fig:mapper_prompt3}
\end{figure*}

All experiments are implemented using the VeRL framework \cite{sheng2024hybridflow}, with Python 3.10 and PyTorch 2.6. We use DeepSeek-R1-0528 \cite{shao2024deepseekmathpushinglimitsmathematical} to generate CoT reasoning traces on the GSM8K \cite{cobbe2021trainingverifierssolvemath} training dataset, guided by prompts that are similar to those in \cite{Liao_He_Xu_Zhang_Liu_Zhao_2025} and shown in Figure~\ref{fig:cot_prompt}. These traces are then distilled in our CoT-Dis setting to supervise SLM training. For STaR, we follow the same setting as DURIT by sampling 8 answers using the SLM and distilling only the correct CoT traces after filtering. For Vanilla-KD \cite{NEURIPS2024_48229913}, we use the mapper model from DURIT as the teacher for knowledge distillation. All distillation methods are trained for 5 epochs, consistent with DURIT Step II. We set the maximum sequence length to 2536, learning rate to $1\text{e}{-5}$, KL loss coefficient to 0.001 and batch size to 4. For GRPO \cite{shao2024deepseekmathpushinglimitsmathematical}, we sample 8 responses per query, and set the batch size to 16, learning rate to $1\text{e}{-6}$, and maximum response length to 1024, and train for 3 epochs. For PRewrite \cite{kong-etal-2024-prewrite}, we perform prompt optimization using the GRPO algorithm, employing DURIT mapper models as translators during optimization. The model is trained for one epoch, consistent with DURIT Step I. For PRewrite, we conduct prompt tuning by evaluating 20 prompt variants—including the original PRewrite prompt—and select the best-performing one, which coincides with the prompt used by the DURIT mapper (as illustrated in Figure~\ref{fig:mapper_prompt}). For DURIT, we adopt the same prompt (see Figure~\ref{fig:mapper_prompt}), with the selection parameter $\lambda$ set to 0.2 for Qwen2.5-0.5B-Instruct model \cite{yang2025qwen3} and 0.05 for Llama3.2-1B-Instruct model \cite{grattafiori2024llama}. For the Qwen2.5-1.5B-Instruct model, we set the self-distillation learning rate in DURIT to 1e-6, and $\lambda$ to 0.01. To assess the robustness of DURIT to prompt variations, we additionally evaluate two alternative prompt designs illustrated in Figures~\ref{fig:mapper_prompt2} and~\ref{fig:mapper_prompt3}. All other training hyperparameters are kept identical to those of the corresponding baselines to ensure fair comparison and set random seed to 1. All models are optimized using the AdamW optimizer with $\beta_1 = 0.9$, $\beta_2 = 0.95$, and weight decay set to 0.01.


\section{Appendix C: Pseudocode of DURIT}
\label{appendix:pseudocode}
We present the pseudocode of DURIT in Algorithm~\ref{alg:durit_pcode}.

\begin{algorithm}[t]
\caption{DURIT: Decoupled Understanding from Reasoning via Iterative Training}
\label{alg:durit_pcode}
\begin{algorithmic}
\Require Dataset $D_0$, Pretrained SLM $R$, Mapper $M$, Codebook $C$
\State Randomly initialize template tokens $\{T_1, \dots, T_n\}$ and query keys $\{k_1, \dots, k_n\}$
\For{iteration = 1 to $N$}
    \Statex \textbf{Step I: Problem Mapper Training}
    \State Cluster $D_0$ into $n$ groups via kNN over problem representations $\mathbf{z}_i$
    \For{each question $Q_i \in D_0$}
        \State Select template token $T_{t_i}$ based on cluster label $t_i$
        \State Construct mapper input $x_i = [Q_i; T_{t_i}]$
        \State Generate mapped question $Q'_i = M(x_i)$
        \State Compute reward $r_i = r_{\text{acc}} + r_{\text{cheating}}$
        \State Compute template similarity loss $\mathcal{L}_{\text{template-sim}}$
        \State Compute key similarity loss $\mathcal{L}_{\text{key-sim}}$
        \State Update $M$ and $C$ using total loss:
        \[
        \mathcal{L}_{\text{total}} = \mathcal{L}_{\text{pg}} + \alpha_1 \mathcal{L}_{\text{key-sim}} + \alpha_2 \mathcal{L}_{\text{template-sim}}
        \]
    \EndFor

    \Statex \textbf{Step II: Self-Distillation Training}
    \State Generate normalized dataset $\mathcal{D}_1 = \{Q_i' = M(Q_i)|Q_i \in D_0\}$
    \For{each $Q_i' \in \mathcal{D}_1$}
        \State Sample $N$ responses $y$ using frozen SLM $R(Q_i')$
        \If{\texttt{answer}$(y_i)$ is correct}
            \State Let $x_s = [Q_i, y_i]$ (student), $x_t = [Q_i', y_i]$ (teacher), and add $(x_s, x_t)$ to $\mathcal{D}_2$
        \EndIf
    \EndFor
    \State Train $R$ on $\mathcal{D}_2$ using the following loss:
    \[
    \mathcal{L}_i = \frac{1}{l} \sum_{k=1}^l \left[(1 - \lambda)(-\log p_s(x_i^k)) + \lambda \, \mathrm{KL}(p_t(x_i^k) \| p_s(x_i^k)) \right]
    \]

    \Statex \textbf{Step III: Reinforcement Learning Training}
    \State Fine-tune $R$ via GRPO on the original dataset $D_0$
\EndFor
\end{algorithmic}
\end{algorithm}

\section{Appendix D: Limitations}
Our proposed approach for decoupling understanding from reasoning offers a novel perspective on enhancing the reasoning capabilities of LLMs. Nevertheless, acknowledging the limitations of our study is also important. Due to hardware constraints, we did not conduct experiments on larger models (e.g., those exceeding 3B parameters). However, the proposed method holds promise for scaling to larger models, as the decoupling mechanism and the dimensionality reduction in the problem space can jointly enhance exploration, learning efficiency, and reasoning robustness. While our approach incurs additional training overhead due to the need for a dedicated mapper, its core strength lies in enhancing reasoning and robustness purely through the model’s own capabilities—without relying on external strong models. Compared to methods like distillation, our approach is more broadly applicable to powerful models, avoiding the high cost of API access and the limitations of requiring a well-matched teacher and large-scale training data. This leads to better scalability and generalization.

\section{Appendix E: Additional Experimental Results}
\label{appendix:more_exp_results}
\subsection{Appendix E.1: Additional Iterative Experiments}
\label{appendix:more_iter}
To investigate the convergence behavior and effectiveness of DURIT's iterative training, we conducted experiments on the GSM8K dataset using the Qwen2.5-0.5B-Instruct and Llama3.2-1B-Instruct models. As shown in Table~\ref{tab:durit-iter}, the average accuracy of both models generally improves with the number of iterations in most cases, confirming the benefits of iterative refinement; however, performance gains gradually diminish due to saturation, typically converging after approximately three iterations, while iterative training on different datasets still exhibits potential for further improvement. To further validate this effect under equal training budgets, we compared two strategies on the Qwen2.5-0.5B-Instruct base model: performing two iterations versus a single iteration with double the training steps (DTT). The results show that, under the same total training time, a single DTT iteration achieves lower accuracy than two iterations, further corroborating the advantage of DURIT's iterative approach.

\begin{table*}[t]
\centering
\small
\setlength{\tabcolsep}{4.5pt}
\renewcommand{\arraystretch}{1.2}
\begin{tabular}{l|c|ccccc|c}
\toprule
\multirow{2}{*}{\textbf{Methods}} & \textbf{In-Domain} & \multicolumn{5}{c|}{\textbf{Out-of-Domain}} & \multirow{2}{*}{\textbf{Average}} \\
& GSM8K-Platinum & MAWPS & SVAMP & MATH500 & GAOKAO & LogiQA &  \\
\midrule
\multicolumn{8}{l}{\textbf{\# Qwen2.5-0.5B-Instruct based}} \\
\midrule
DURIT (iter=1) & 53.68 & 60.19 & 62.67 & 31.00 & \underline{23.39} & 24.58 & 42.59 \\
DURIT (iter=1, DTT) & \textbf{54.43} & \underline{60.58} & 60.00 & \underline{31.40} & \textbf{24.19} & 25.19 & 42.63 \\
DURIT (iter=2) & 53.10 & 60.38 & 63.00 & \textbf{32.80} & 22.58 & \underline{25.81} & \underline{42.95} \\
DURIT (iter=3) & 53.76 & \underline{60.58} & \underline{63.67} & 31.00 & \textbf{24.19} & 25.65 & \textbf{43.14} \\
DURIT (iter=4) & \underline{54.34} & \textbf{61.15} & \textbf{64.33} & 31.00 & 21.77 & \textbf{26.27} & \textbf{43.14} \\
\midrule
\multicolumn{8}{l}{\textbf{\# Llama3.2-1B-Instruct based}} \\
\midrule
DURIT (iter=1) & 50.37 & 59.62 & 64.33 & 26.00 & \textbf{14.52} & \textbf{21.20} & 39.34 \\
DURIT (iter=2) & 52.36 & \textbf{62.31} & 66.00 & \underline{27.60} & 12.10 & \underline{19.82} & \underline{40.03} \\
DURIT (iter=3) & \underline{52.85} & 61.54 & \textbf{68.00} & \textbf{27.80} & \underline{13.71} & 15.67 & 39.93 \\
DURIT (iter=4) & \textbf{53.43} & \underline{61.76} & \underline{67.00} & 26.60 & \textbf{14.52} & 17.05 & \textbf{40.06} \\
\bottomrule
\end{tabular}
\caption{Iterative performance (\%) of DURIT on six benchmarks using \textbf{Qwen2.5-0.5B-Instruct} and \textbf{Llama3.2-1B-Instruct} as base models. \textbf{Bold} and \underline{underline} indicate the best and second-best results, respectively. DTT denotes double training time for a single iteration.}
\label{tab:durit-iter}
\end{table*}

\subsection{Appendix E.2: Additional Model Evaluation}
\label{appendix:more_model_eval}
To further validate the effectiveness of our approach, we conduct experiments on the Qwen2.5-1.5B-Instruct model, with results presented in Table~\ref{tab:appendix_acc_results}. After a single iteration, DURIT achieves an average accuracy improvement of 1.01\% over the strongest extreme baselines, with particularly notable gains on the MATH500 and GAOKAO datasets, reaching 53.60\% and 32.26\%, respectively. A second iteration yields further improvements, demonstrating the scalability and stability of the method. While the GSM8K dataset offers limited additional benefit due to its relative simplicity, DURIT exhibits greater potential when trained on more complex datasets, highlighting its ability to enhance exploration efficiency by explicitly decoupling understanding from reasoning. Besides, as shown in Table~\ref{tab:appendix_robost_results}, DURIT shows strong robustness to reasoning: it achieves even higher accuracy on the gsm-symbolic \cite{mirzadeh2024gsmsymbolicunderstandinglimitationsmathematical} dataset than on the original GSM8K-100 subset and outperforms all extreme baselines in terms of reduced performance drop under symbolic perturbations. These results confirm that DURIT effectively mitigates the influence of spurious correlations by performing reasoning in a more abstract and canonical problem space, leading to consistent improvements in both reasoning accuracy and robustness.

\begin{table*}[t]
\centering
\small
\setlength{\tabcolsep}{4.5pt}
\renewcommand{\arraystretch}{1.2}
\begin{tabular}{l|c|ccccc|c}
\toprule
\multirow{2}{*}{\textbf{Methods}} & \textbf{In-Domain} & \multicolumn{5}{c|}{\textbf{Out-of-Domain}} & \multirow{2}{*}{\textbf{Average}} \\
& gsm8k-platinum & MAWPS & SVAMP & MATH500 & GAOKAO & LogiQA & \\
\midrule
\multicolumn{8}{l}{\textbf{\# Qwen2.5-1.5B-Instruct based}} \\
\midrule
Base \cite{yang2025qwen3} & 72.46 & 57.88 & 73.00 & 46.40 & 25.00 & 40.86 & 52.60 \\
CoT-Dis \cite{magister-etal-2023-teaching} & 71.63 & 59.42 & \textbf{85.67} & 42.40 & 20.97 & 37.48 & 52.93 \\
STaR \cite{zelikman2024star} & \underline{76.34} & 63.27 & 80.67 & 49.40 & 26.61 & \underline{42.24} & 56.42 \\
GRPO \cite{shao2024deepseekmathpushinglimitsmathematical} & \textbf{76.43} & \textbf{63.85} & 83.67 & 50.20 & 28.23 & 39.32 & 56.95 \\
PRewrite \cite{kong-etal-2024-prewrite} & 70.55 & 54.42 & 80.00 & 45.60 & 20.16 & 31.03 & 50.29 \\
Vanilla-KD \cite{NEURIPS2024_48229913} & 76.18 & 62.69 & 82.00 & 51.80 & 24.19 & 36.56 & 55.57 \\
DURIT (ours, iter=1) & \underline{76.34} & \underline{63.46} & 81.67 & \textbf{53.60} & \textbf{32.26} & 40.40 & \underline{57.96} \\
DURIT (ours, iter=2) & 76.01 & 62.88 & \underline{84.00} & \underline{52.20} & \underline{30.65} & \textbf{42.40} & \textbf{58.02} \\
\bottomrule
\end{tabular}
\caption{Performance (\%) of the Qwen2.5-1.5B-Instruct model trained on the GSM8K dataset, evaluated across six representative benchmarks. The \textbf{bold} and \underline{underline} indicate the best and second-best results, respectively.}
\label{tab:appendix_acc_results}
\end{table*}

\begin{table}[ht]
\label{tab:qwen15-results}
\centering
\begin{tabular}{l|ccc}
\toprule
\multirow{2}{*}{Method} & \multicolumn{3}{c}{Qwen-1.5B} \\
\cmidrule{2-4}
 & Orig & Symb & $\Delta\%$ \\
\midrule
Base & 67.00    & 63.82    & -4.75 \\
CoT-Dis & 71.00    & 68.24    & -3.89 \\
STaR & 79.00    & 67.10    & -15.06 \\
GRPO & 65.00    & 66.50    & \underline{+2.31} \\
PRewrite & 70.00    & 63.82    & -8.83 \\
Vanilla-KD & 73.00    & 68.34    & -6.38 \\
DURIT & 64.00 & 66.68 & \textbf{+4.19} \\
\bottomrule
\end{tabular}
\caption{Comparison of different methods based on Qwen2.5-1.5B-Instruct trained on the GSM8K dataset. DURIT is trained with a single iteration. Orig: original GSM8K-100 subset; Symb: gsm-symbolic; $\Delta\%$: relative performance drop from Orig to Symb. \textbf{Bold} and \underline{underline} indicate best and second-best results in each group.}
\label{tab:appendix_robost_results}
\end{table}

\subsection{Appendix E.3: Additional Dataset Evaluation}
To evaluate the generalizability of DURIT beyond mathematical domains, we conducted additional training using the Qwen2.5-1.5B-Instruct model on the logical reasoning dataset LogiQA. Since LogiQA consists of multiple-choice questions, we introduced a subset of fill-in-the-blank samples from GSM8K to mitigate potential format-specific forgetting and preserve the model’s ability to generalize across answer formats. Specifically, we constructed a mixed training set of 6,000 examples by randomly sampling 4,000 instances from LogiQA and 2,000 from GSM8K. As shown in Table~\ref{tab:appendix_dataset_acc_results}, the CoT-Dis method based on DeepSeek-R1 performs notably worse, primarily due to the overly complex CoT rationales generated on the LogiQA dataset. These complex reasoning traces are difficult for SLMs to learn and generalize from, ultimately impairing the student model’s reasoning capability across tasks. This observation reveals a potential limitation of traditional knowledge distillation approaches when the teacher outputs are misaligned with the student’s learning capacity. In contrast, DURIT achieves the best overall accuracy, despite being primarily trained on logical reasoning data. It surpasses the strongest baseline by +0.43\% and +0.62\% on average, and +1.23\% on LogiQA dataset after one and two iterations, respectively, and consistently delivers high accuracy across most benchmarks, including those focused on mathematical reasoning.  Moreover, as presented in Table~\ref{tab:appendix_dataset_robost_results}, DURIT demonstrates superior robustness, even achieving higher accuracy on the gsm-symbolic dataset, which highlights its enhanced reasoning ability and robustness via the proposed decoupling of understanding and reasoning. These results confirm the effectiveness and domain transferability of DURIT in improving general reasoning performance.

\begin{table*}[t]
\centering
\small
\setlength{\tabcolsep}{4.5pt}
\renewcommand{\arraystretch}{1.2}
\begin{tabular}{l|cc|cccc|c}
\toprule
\multirow{2}{*}{\textbf{Methods}} 
& \multicolumn{2}{c|}{\textbf{In-Domain}} 
& \multicolumn{4}{c|}{\textbf{Out-of-Domain}} 
& \multirow{2}{*}{\textbf{Average}} \\
& gsm8k-platinum & LogiQA & SVAMP & MATH500 & GAOKAO & MAWPS & \\
\midrule
\multicolumn{8}{l}{\textbf{\# Qwen2.5-1.5B-Instruct based}} \\
\midrule
Base \cite{yang2025qwen3} & 72.46 & 40.86 & 73.00 & 46.40 & 25.00 & 57.88 & 52.60 \\
CoT-Dis \cite{magister-etal-2023-teaching} & 69.73 & 30.26 & 74.33 & 33.20 & 21.77 & 55.58 & 47.48 \\
STaR \cite{zelikman2024star} & 74.44 & \underline{43.93} & \underline{84.00} & 49.40 & 25.00 & 62.69 & 56.58 \\
GRPO \cite{shao2024deepseekmathpushinglimitsmathematical} & 75.85 & 43.63 & 83.67 & 50.20 & \underline{26.61} & 62.69 & 57.11 \\
PRewrite \cite{kong-etal-2024-prewrite} & 70.14 & 31.64 & 79.00 & 46.60 & 23.39 & 55.38 & 51.03 \\
Vanilla-KD \cite{NEURIPS2024_48229913} & \underline{75.93} & 39.01 & 82.00 & \underline{51.60} & \textbf{27.42} & \textbf{63.65} & 56.60 \\
DURIT (ours, iter=1) & \textbf{76.84} & 43.47 & \underline{84.00} & 50.60 & \textbf{27.42} & \underline{62.88} & \underline{57.54} \\
DURIT (ours, iter=2) & 75.43 & \textbf{45.16} & \textbf{85.00} & \textbf{53.00} & \underline{26.61} & 61.15 & \textbf{57.73} \\
\bottomrule
\end{tabular}
\caption{Performance (\%) of the Qwen2.5-1.5B-Instruct model trained on the GSM8K + LogiQA mixed dataset, evaluated across six representative benchmarks. The \textbf{bold} and \underline{underline} indicate the best and second-best results, respectively.}
\label{tab:appendix_dataset_acc_results}
\end{table*}

\begin{table}[ht]
\label{tab:qwen15-results}
\centering
\begin{tabular}{l|ccc}
\toprule
\multirow{2}{*}{Method} & \multicolumn{3}{c}{Qwen-1.5B} \\
\cmidrule{2-4}
 & Orig & Symb & $\Delta\%$ \\
\midrule
Base & 67.00    & 63.82    & \underline{-4.75} \\
CoT-Dis & 70.00    & 55.20    & -21.14 \\
STaR & 73.00    & 66.70    & -8.63 \\
GRPO & 70.00    & 66.04    & -5.66 \\
PRewrite & 68.00    & 63.60    & -6.47 \\
Vanilla-KD & 76.00    & 68.42    & -9.97 \\
DURIT & 64.00 & 65.42 & \textbf{+2.22} \\
\bottomrule
\end{tabular}
\caption{Comparison of different methods based on Qwen2.5-1.5B-Instruct trained on the GSM8K + LogiQA mixed dataset. DURIT is trained with a single iteration. Orig: original GSM8K-100 subset; Symb: gsm-symbolic; $\Delta\%$: relative performance drop from Orig to Symb. \textbf{Bold} and \underline{underline} indicate best and second-best results in each group.}
\label{tab:appendix_dataset_robost_results}
\end{table}

\subsection{Appendix E.4: Parameter Analysis}
\begin{figure}[htp]
    \centering
    \includegraphics[width=1\columnwidth]{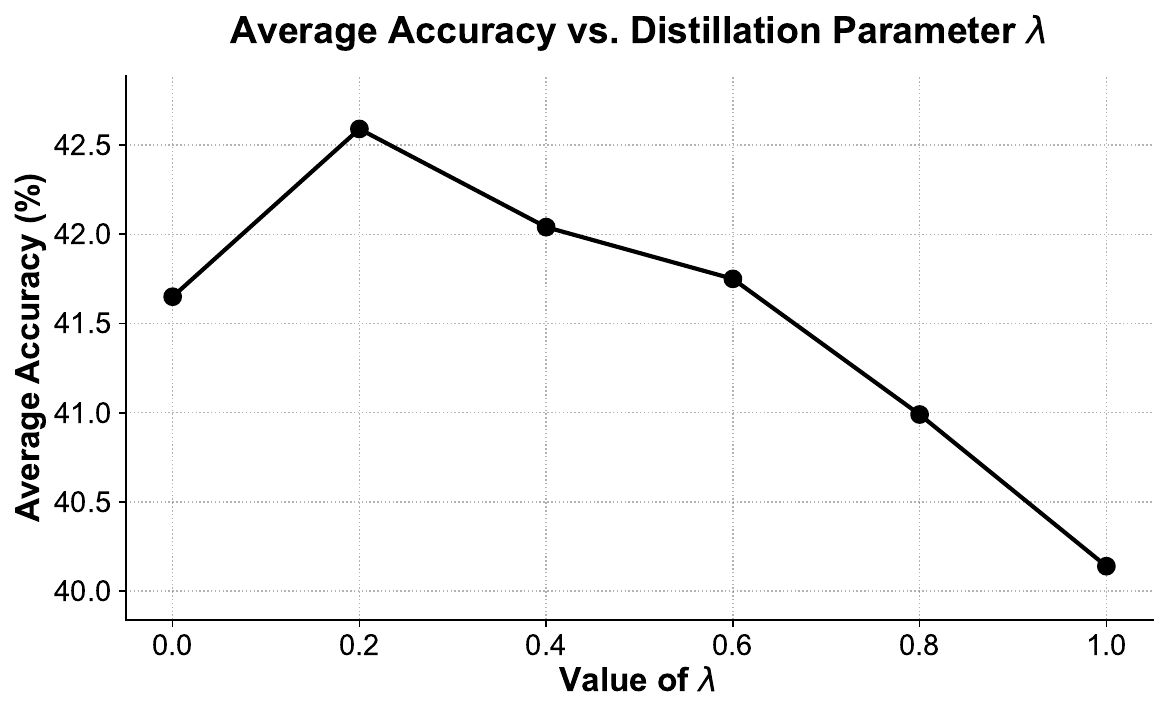} 
    \caption{Average accuracy of Qwen2.5-0.5B-Instruct across six datasets under different distillation coefficient $\lambda$.}
    \label{fig:lambda_analysis}
\end{figure}

To investigate the impact of hyperparameters on the algorithm’s performance, we conduct an in-depth analysis of the distillation coefficient $\lambda$ using the Qwen2.5-0.5B-Instruct model. The experimental results are illustrated in Figure~\ref{fig:lambda_analysis}. We observe that when $\lambda$ is too small, the training objective is dominated by the supervised fine-tuning (SFT) loss, which may overly restrict the model’s capacity for exploration during the reinforcement learning (RL) phase. As $\lambda$ increases, performance improves and reaches its peak at $\lambda = 0.2$. However, further increasing $\lambda$ leads to a performance drop, likely due to the excessive influence of self-distillation. In this case, the model may become overly reliant on potentially inaccurate interpretations of the transformed question \( Q' \), introducing harmful noise that not only degrades learning but may also disrupt the model's original internal reasoning patterns. Overall, the performance remains relatively stable across a broad range of $\lambda$ values (from 0.1 to 0.5), with $\lambda = 0.2$ emerging as the most effective choice.

\begin{figure}[htp]
    \centering
    \includegraphics[width=1\columnwidth]{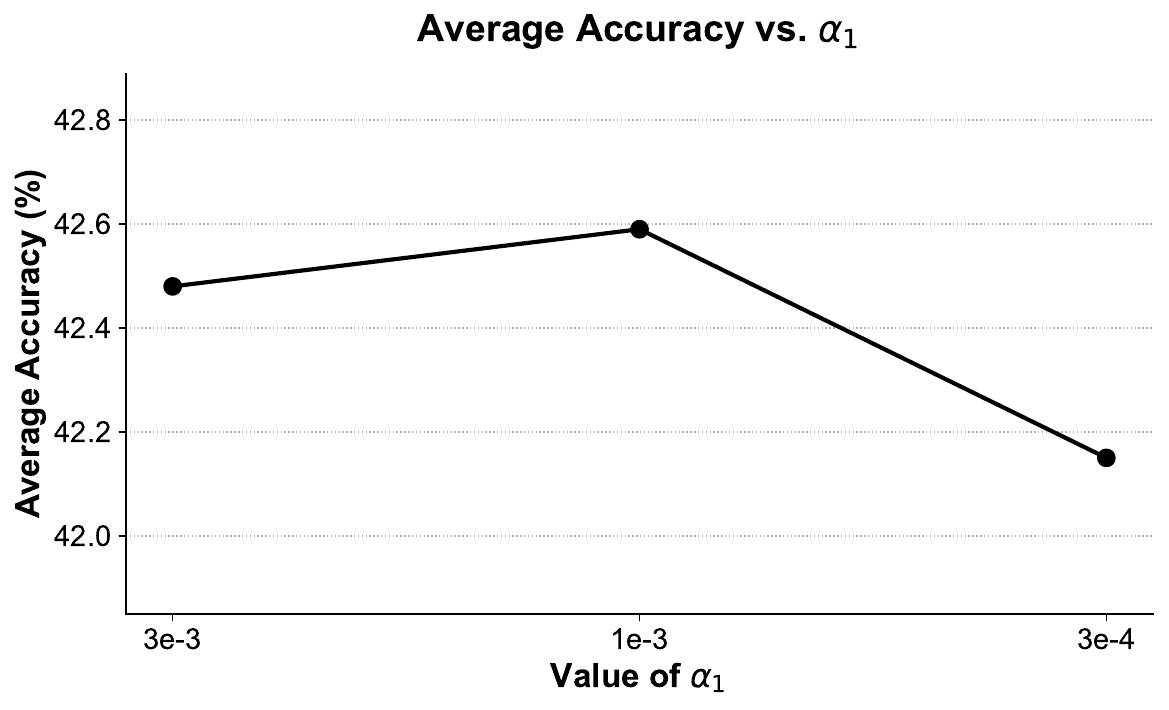} 
    \caption{Average accuracy of Qwen2.5-0.5B-Instruct across six datasets under different distillation coefficient $\alpha_1$.}
    \label{fig:alpha1_analysis}
\end{figure}

\begin{figure}[htp]
    \centering
    \includegraphics[width=1\columnwidth]{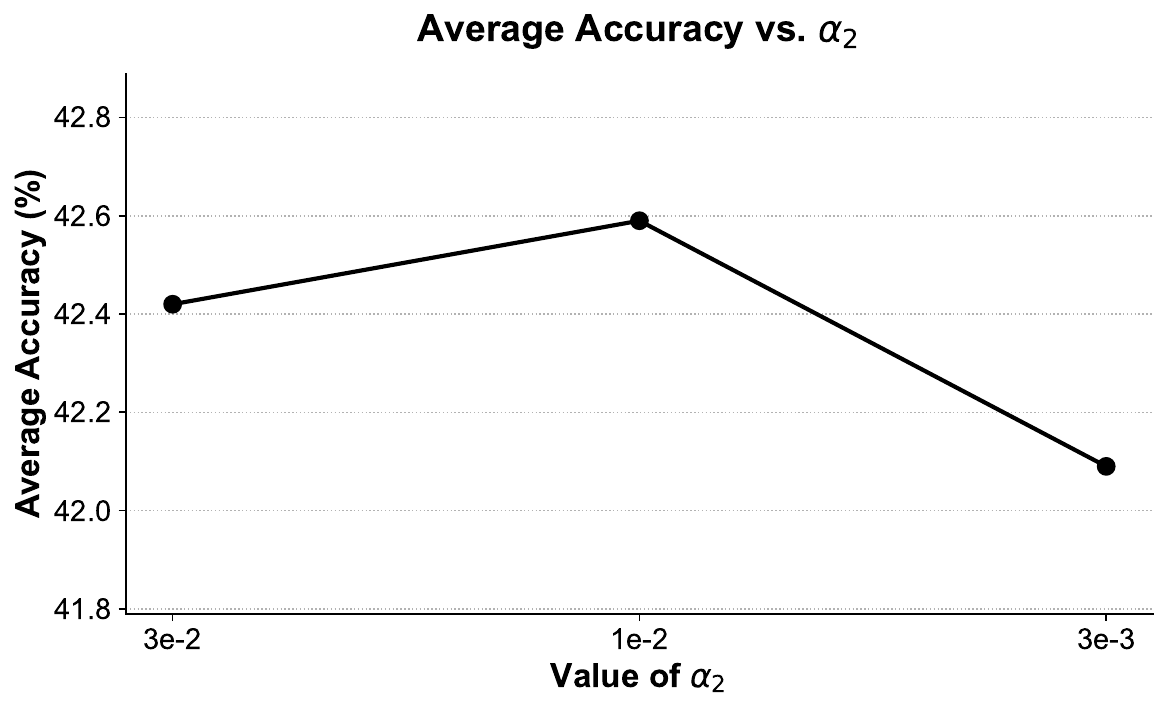} 
    \caption{Average accuracy of Qwen2.5-0.5B-Instruct across six datasets under different distillation coefficient $\alpha_2$.}
    \label{fig:alpha2_analysis}
\end{figure}

We also analyzed the sensitivity of $\alpha_1$ and $\alpha_2$. As shown in Figures~\ref{fig:alpha1_analysis} and~\ref{fig:alpha2_analysis}, across a threefold variation in each parameter, the model's accuracy fluctuates by no more than 0.5\%, demonstrating that DURIT exhibits substantial robustness to reasonable parameter choices. The best performance is achieved when $\alpha_1$=1e-3 and $\alpha_2$=1e-2. Accordingly, we adopt these values as the default settings.

\subsection{Appendix E.5: Training Reward Comparison}
\begin{figure}[htp]
    \centering
    \includegraphics[width=1\columnwidth]{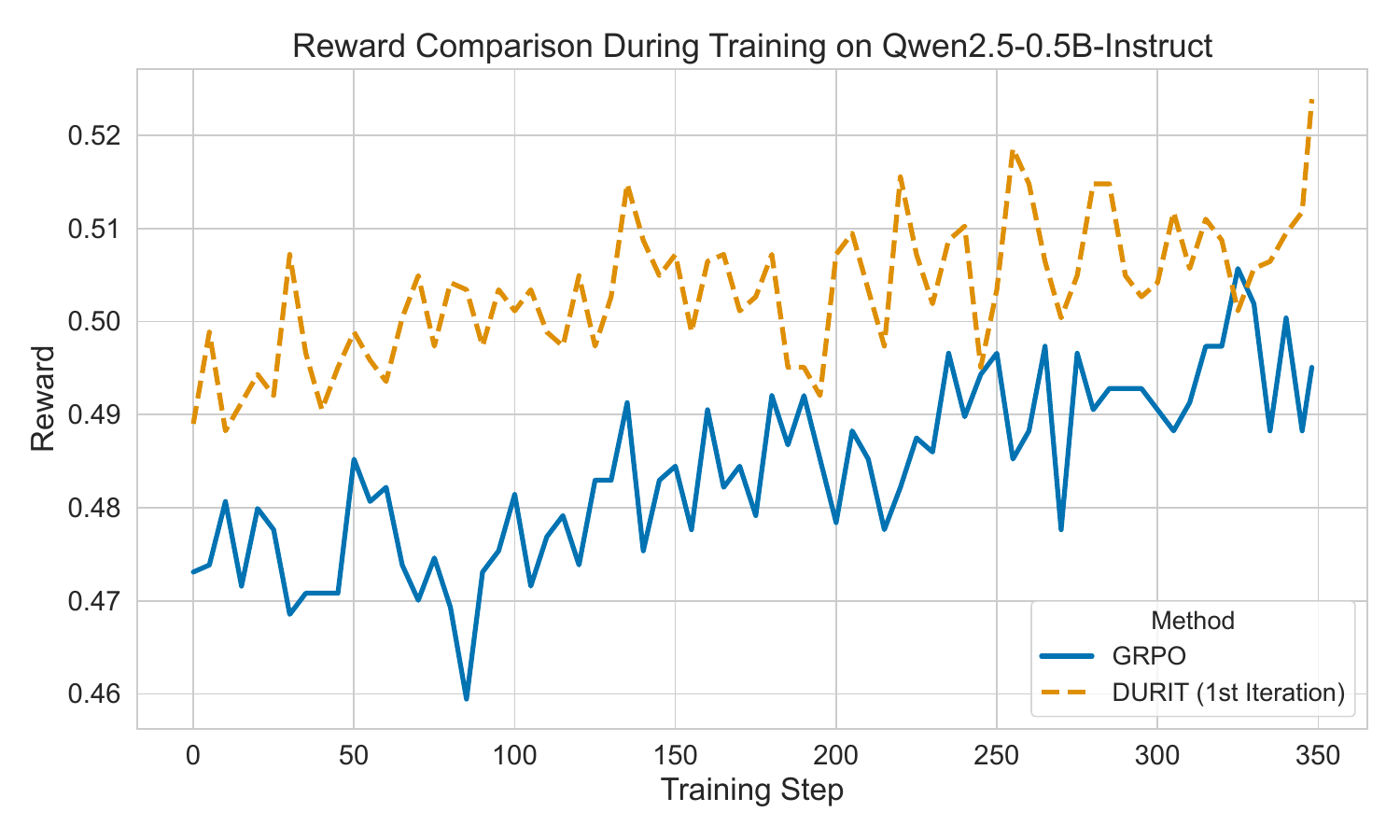} 
    \caption{Training reward comparison of Qwen2.5-0.5B-Instruct between DURIT-step III and the GRPO baseline.}
    \label{fig:qwen0.5b_reward}
\end{figure}

\begin{figure}[htp]
    \centering
    \includegraphics[width=1\columnwidth]{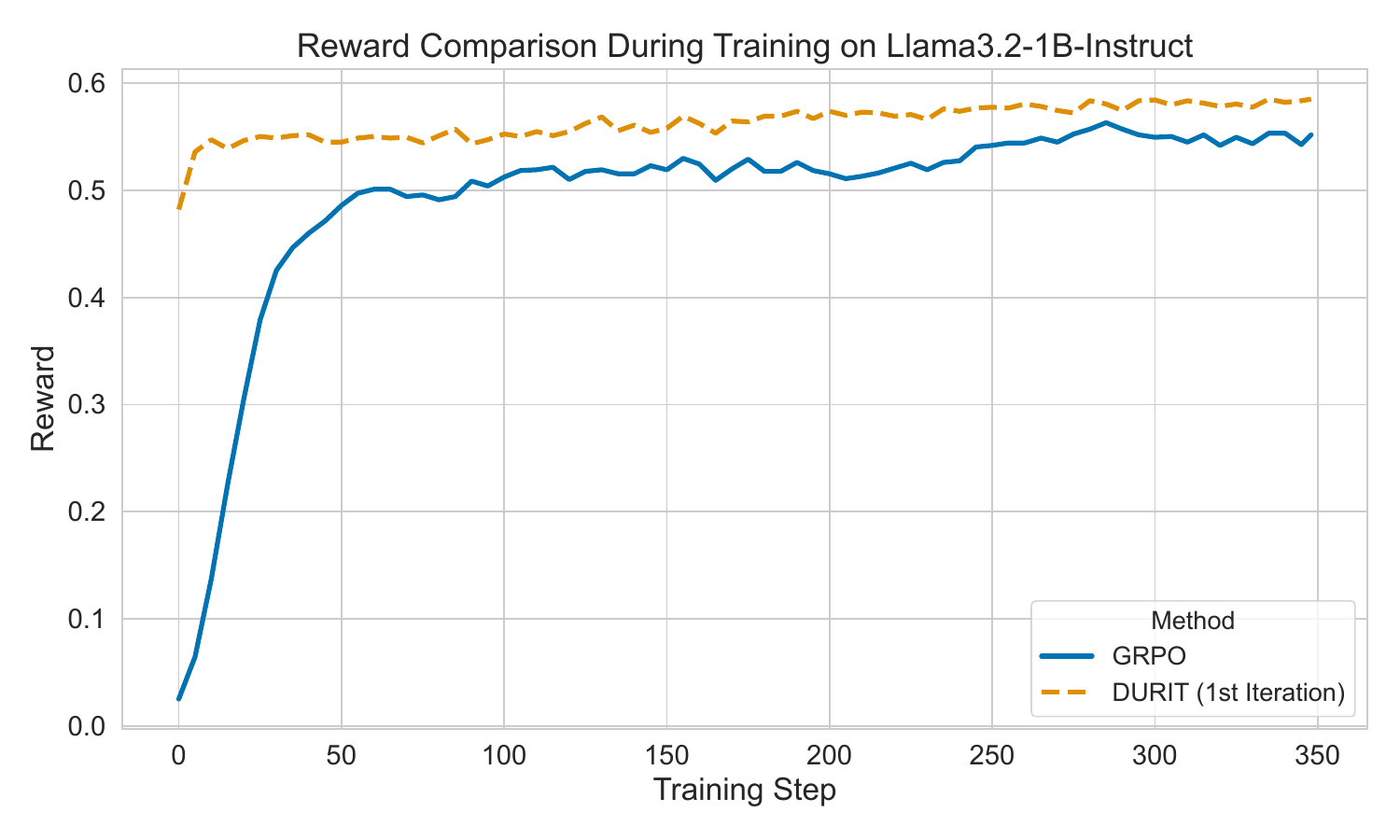} 
    \caption{Training reward comparison of Llama3.2-1B-Instruct between DURIT-step III and the GRPO baseline.}
    \label{fig:lambda1b_reward}
\end{figure}

To further validate DURIT’s effectiveness in improving training efficiency, we compare the training rewards of DURIT (Step III of the first iteration) with those of the GRPO baseline on the Qwen2.5-0.5B-Instruct and Llama3.2-1B-Instruct models, as shown in Figures~\ref{fig:qwen0.5b_reward} and~\ref{fig:lambda1b_reward}. For both models, DURIT achieves the same reward level as the converged GRPO baseline after only a fraction of the training steps, significantly reducing training time and improving sample efficiency. Moreover, DURIT yields a higher final reward upon convergence, demonstrating its ability not only to accelerate reinforcement learning by decoupling understanding from reasoning and compressing the effective state space, but also to improve the quality of convergence. These results also provide empirical support for Theorem~\ref{appendix:theorem1}.


\subsection{Appendix E.6: Comparison of Computational Cost}
\label{appendix:training_overhead}
To compare the training overhead of all methods, we report the total runtime measured on a single NVIDIA A800 GPU with 80GB memory, using Qwen2.5-0.5B-Instruct as the base model, as summarized in Table~\ref{appendix:time-comparison}. For CoT-Dis, the primary bottleneck lies in collecting CoT data from a strong teacher model. In the case of DeepSeek-R1, due to relatively slow API responses, we employed three concurrent terminals but still required 60 hours to complete data collection, in addition to incurring extra API usage costs. For STaR, generating CoT outputs from the base model takes approximately 2 hours, resulting in a total runtime of 3.2 hours. For GRPO, training converges within 3 epochs, taking a total of 6.2 hours. PRewrite converges within just 0.23 epochs of training (1.7h); however, it introduces additional inference-time overhead due to the requirement of a question rewriter as a preprocessing step. Vanilla-KD spends 2.4 hours collecting CoT responses from the teacher model, and completes training in 2.3 hours, totaling 4.7 hours. For DURIT, Step I trains the problem-space mapper to convergence in 1.7 hours (0.23 epochs). Step II performs question mapping and collects base model CoT responses in 2.8 hours, followed by 1.2 hours of self-distillation. Step III involves GRPO training for 6.2 hours, resulting in a total runtime of 11.9 hours. Notably, DURIT without Step III (i.e., w/o grpo) already achieves state-of-the-art performance, as shown in the ablation results in the main paper, with a total training cost of only 5.7 hours. Furthermore, when Step III reinforcement learning is trained on just 10\% of the data, the model surpasses the full GRPO baseline in terms of reward (see Figure~\ref{fig:qwen0.5b_reward}), with total training time only 0.1 hours longer than GRPO. These results highlight that DURIT, by decoupling understanding from reasoning and effectively compressing the problem space, improves both exploration efficiency and convergence speed. This offers a new perspective for enhancing the reasoning capability and robustness of LLMs.

\begin{table}[t]
  \setlength{\tabcolsep}{4pt}
  \centering
  \begin{tabular}{lccc}
    \toprule
    Method & Data Processing & Training & Total Time \\
    \midrule
    CoT-Dis      & 60.0  & 1.6  & 61.6 \\
    STaR         & 2.0  & 1.2  & 3.2 \\
    GRPO         & 0  & 6.2  & 6.2 \\
    PRewrite     & 0  & 1.7  & 1.7 \\
    Vanilla-KD   & 2.4  & 2.3  & 4.7 \\
    DURIT        & 2.8  & 9.1  & 11.9 \\
    DURIT w/o grpo & 2.8  & 2.9  & 5.7 \\
    DURIT grpo 10\%      & 2.8  & 3.5  & 6.3 \\
    \bottomrule
  \end{tabular}
  \caption{Estimated computational time (in hours) for different methods using the Qwen2.5-0.5B-Instruct model.}
  \label{appendix:time-comparison}
\end{table}

\section{Appendix F: Case Study}
\label{appendix:case_study}
\begin{figure*}[htp]
    \centering
    \includegraphics[width=2\columnwidth]{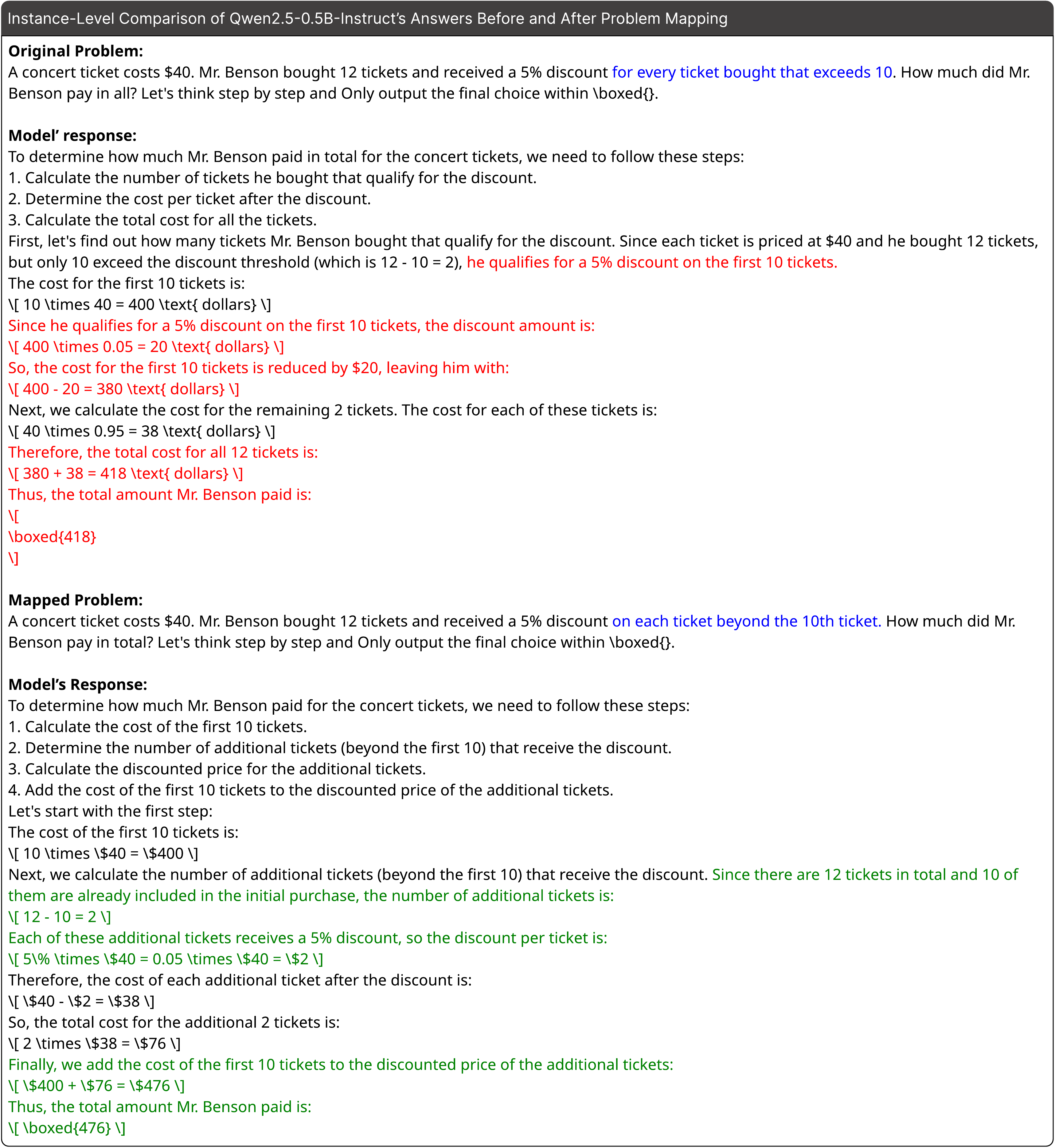} 
    \caption{Illustrative comparison of Qwen2.5-0.5B-Instruct’s responses before and after problem-space mapping, with 
    \textcolor{red}{red} highlighting incorrect reasoning, \textcolor{green!60!black}{green} indicating correct reasoning, and 
    \textcolor{blue}{blue} marking key differences.}
    \label{fig:qwen05b_response_case1}
\end{figure*}

\begin{figure*}[htp]
    \centering
    \includegraphics[width=2\columnwidth]{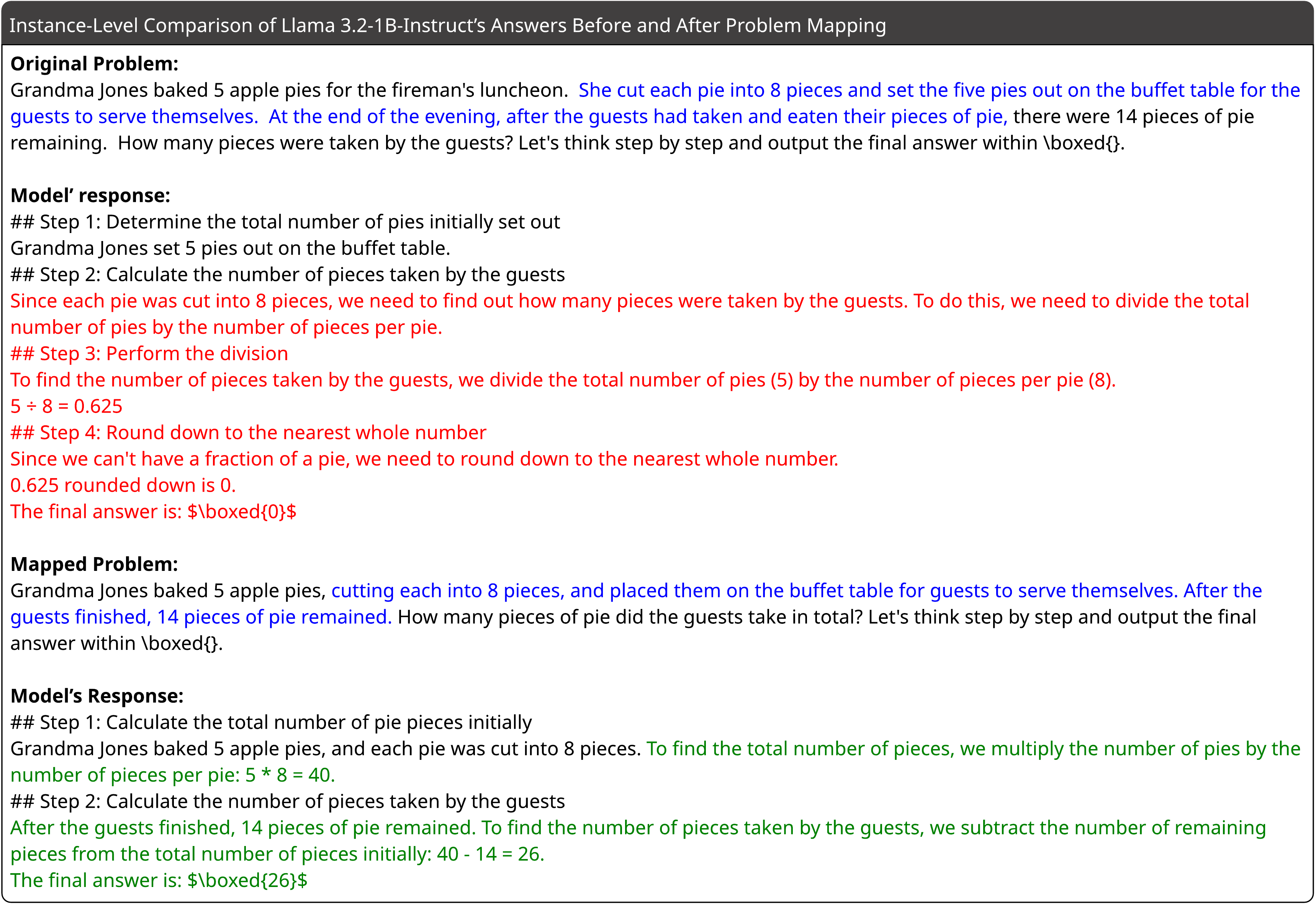} 
    \caption{Illustrative comparison of Llama 3.2-1B-Instruct’s responses before and after problem-space mapping, with 
    \textcolor{red}{red} highlighting incorrect reasoning, \textcolor{green!60!black}{green} indicating correct reasoning, and 
    \textcolor{blue}{blue} marking key differences.}
    \label{fig:llama1b_response_case2}
\end{figure*}

Illustrative examples of the original and mapped questions, along with the corresponding responses from the Qwen2.5-0.5B-Instruct and Llama 3.2-1B-Instruct base models, are presented in Figure~\ref{fig:qwen05b_response_case1} and Figure~\ref{fig:llama1b_response_case2}, respectively. Prior to mapping, the model misinterpreted the discount mechanism described in the original question, resulting in an incorrect answer. After transformation by the problem space mapper $M$, the question was reformulated in a more concise and standardized manner, effectively reducing the model's comprehension burden. This enabled more focused reasoning, ultimately leading to the correct solution.

\begin{figure*}[htp]
    \centering
    \includegraphics[width=2\columnwidth]{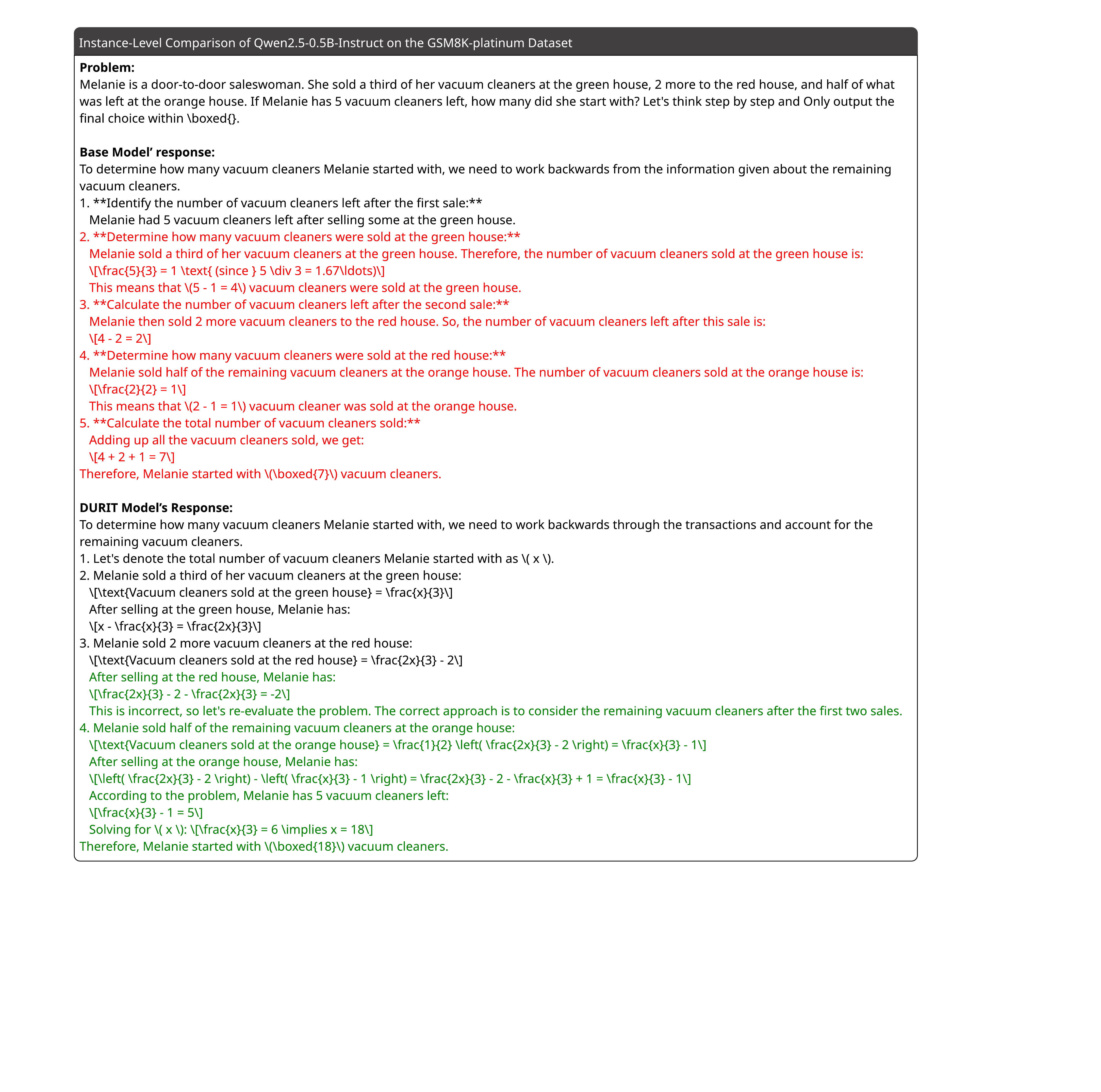} 
    \caption{Instance-level comparison of Qwen2.5-0.5B-Instruct on the GSM8K-platinum dataset, contrasting responses from the base model and DRUIT model, with \textcolor{red}{red} indicating incorrect reasoning and \textcolor{green!60!black}{green} denoting correct reasoning.}
    \label{fig:qwen05b_gsm8k_platinum_case}
\end{figure*}

\begin{figure*}[htp]
    \centering
    \includegraphics[width=2\columnwidth]{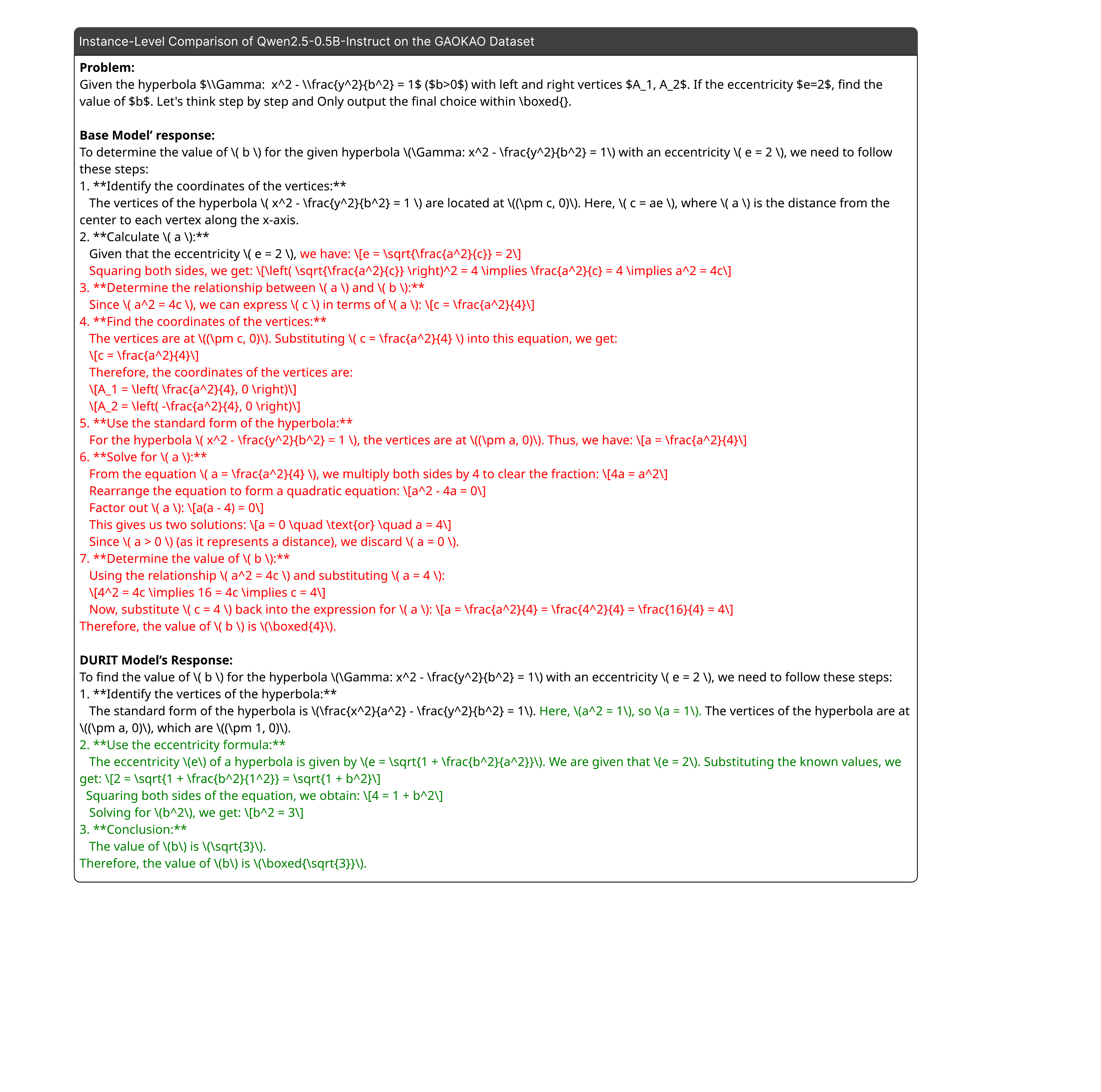} 
    \caption{Instance-level comparison of Qwen2.5-0.5B-Instruct on the GAOKAO dataset, contrasting responses from the base model and DRUIT model, with \textcolor{red}{red} indicating incorrect reasoning and \textcolor{green!60!black}{green} denoting correct reasoning.}
    \label{fig:qwen05b_gaokao_case}
\end{figure*}

\begin{figure*}[htp]
    \centering
    \includegraphics[width=2\columnwidth]{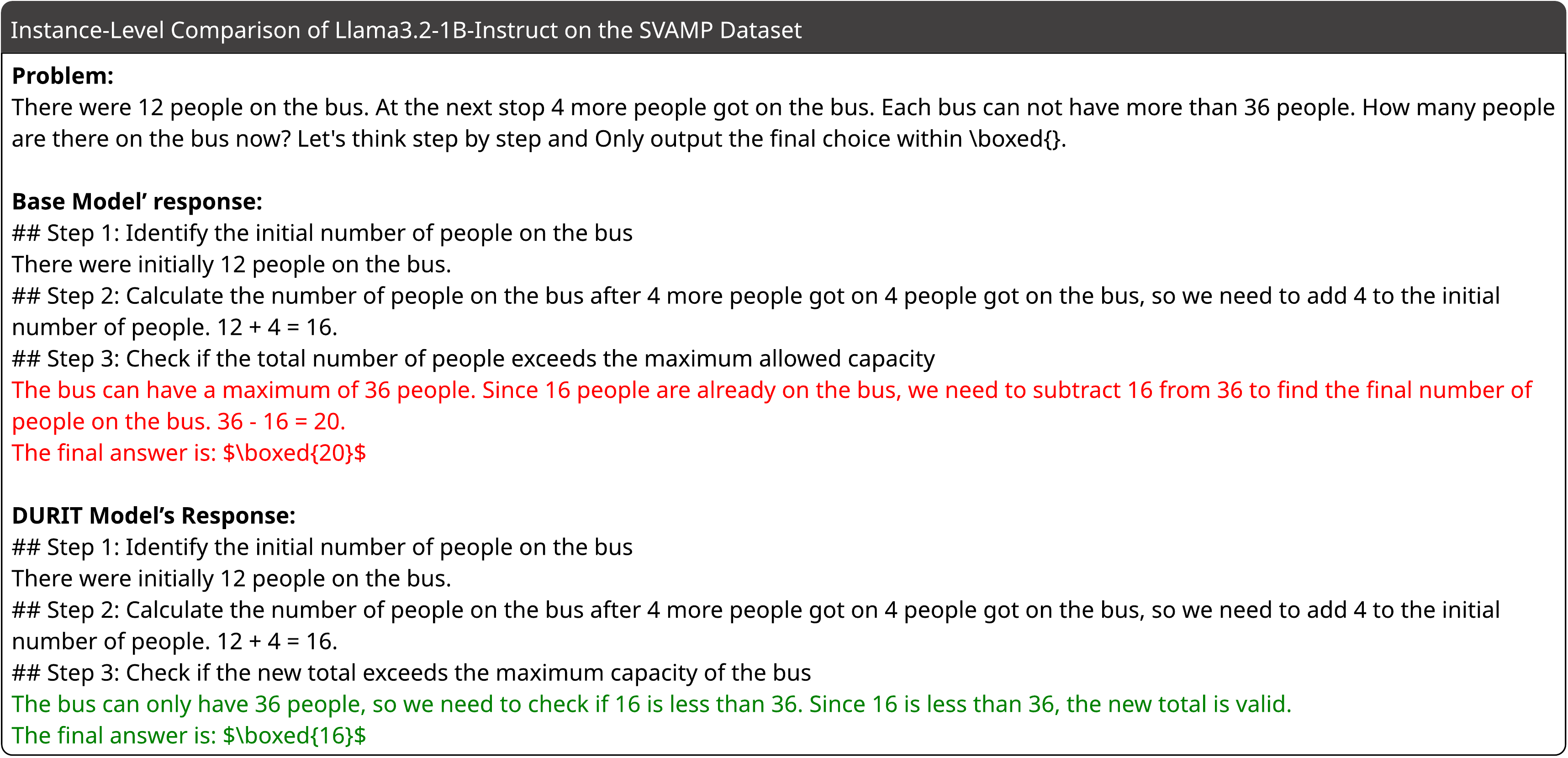} 
    \caption{Instance-level comparison of Llama3.2-1B-Instruct on the SVAMP dataset, contrasting responses from the base model and DRUIT model, with \textcolor{red}{red} indicating incorrect reasoning and \textcolor{green!60!black}{green} denoting correct reasoning.}
    \label{fig:llama1b_svamp_case}
\end{figure*}

\begin{figure*}[htp]
    \centering
    \includegraphics[width=2\columnwidth]{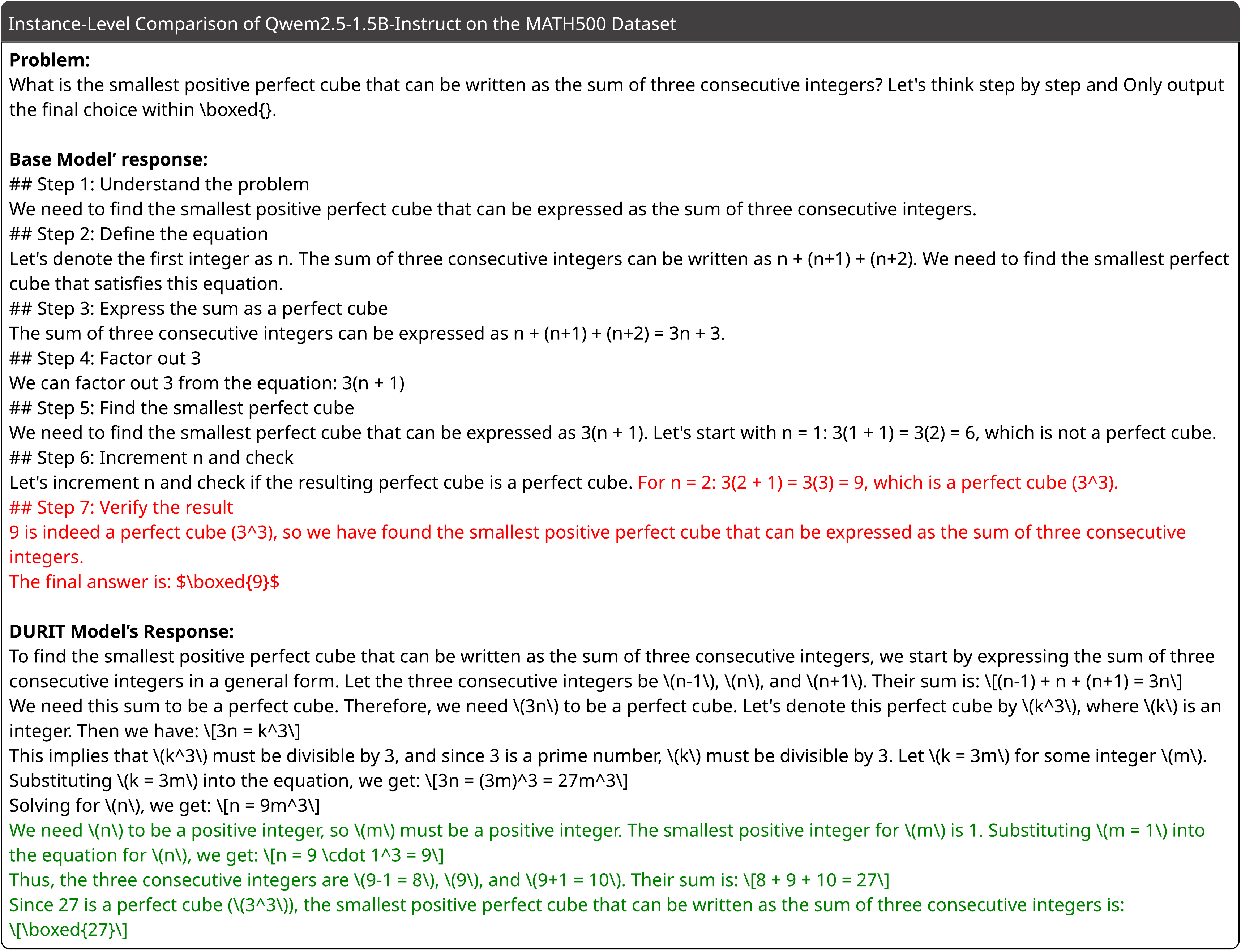} 
    \caption{Instance-level comparison of Qwen2.5-1.5B-Instruct on the MATH500 dataset, contrasting responses from the base model and DRUIT model, with \textcolor{red}{red} indicating incorrect reasoning and \textcolor{green!60!black}{green} denoting correct reasoning.}
    \label{fig:qwen15b_math500_case}
\end{figure*}

Figure \ref{fig:qwen05b_gsm8k_platinum_case}, Figure \ref{fig:qwen05b_gaokao_case}, Figure \ref{fig:llama1b_svamp_case} and Figure \ref{fig:qwen15b_math500_case} provide illustrative examples from the GSM8K-Platinum \cite{vendrow2025largelanguagemodelbenchmarks}, GAOKAO \cite{zhang2024evaluatingperformancelargelanguage}, SVAMP \cite{patel2021nlpmodelsreallyable} and MATH500 \cite{hendrycks2021measuringmathematicalproblemsolving} datasets, highlighting the significant reasoning enhancements conferred by our DURIT methodology. In the first case (Figure \ref{fig:qwen05b_gsm8k_platinum_case}), the base model, confused by complex problem details, utilizes a flawed backward-reasoning approach. Conversely, the DURIT-trained model correctly interprets the problem, adopts a robust equation-solving strategy, and leverages a self-correction mechanism—activated by RL—to rectify errors mid-solution and achieve the correct outcome. In the second case (Figure \ref{fig:qwen05b_gaokao_case}), the base model's reliance on an incorrect formula leads to a verbose and erroneous answer. The DURIT-trained model, however, demonstrates superior reasoning, delivering a solution that is both concise and correct. In the third case (Figure~\ref{fig:llama1b_svamp_case}), the base model is misled by the distracting information regarding the bus capacity, resulting in an incorrect calculation. In contrast, the DURIT-trained model accurately interprets the constraint imposed by the bus capacity, performs the correct conditional reasoning, and arrives at the correct answer. In the fourth case (Figure~\ref{fig:qwen15b_math500_case}), the base model incorrectly identifies 9 as the cube of 3, leading to an erroneous conclusion. In contrast, the DURIT-trained model adopts a correct and more concise reasoning path, arriving at the correct answer.

\end{document}